\ifdefined\pdfoutput
  \pdfoutput=1
\fi
\documentclass[10pt,journal,compsoc]{IEEEtran}

\usepackage[T1]{fontenc}
\usepackage[utf8]{inputenc}
\usepackage{times}
\usepackage{microtype}
\usepackage{inconsolata}
\usepackage{marvosym}
\usepackage{xcolor}
\usepackage{amsmath}
\usepackage{array}
\usepackage{booktabs}
\usepackage{multirow}
\usepackage{makecell}
\usepackage{graphicx}
\usepackage{stfloats}
\usepackage{cite}
\usepackage{url}
\usepackage{enumitem}
\usepackage[usestackEOL]{stackengine}
\usepackage{xspace}
\usepackage{pifont}
\usepackage{hyperref}
\hypersetup{
  hidelinks,
  pdftitle={Open Evaluation Agent: Efficient and Promptable Evaluation of Visual Generative Models},
  pdfauthor={Shulin Tian, Ziqi Huang, Fan Zhang, Hongyuan Zhu, Yu Qiao, and Ziwei Liu},
  pdfsubject={Efficient and promptable evaluation of visual generative models},
  pdfkeywords={evaluation agent, visual generation, text-to-image, text-to-video}
}

\newcommand{\cmark}{\textcolor[rgb]{0.13,0.55,0.13}{\ding{51}}}
\newcommand{\xmark}{\textcolor[rgb]{0.5,0,0}{\ding{55}}}
\newcommand{\ie}{\textit{i.e.}\xspace}

\begin{document}

\title{Open Evaluation Agent: Efficient and Promptable Evaluation of Visual Generative Models}


\author{Shulin Tian, Ziqi Huang, Fan Zhang, Hongyuan Zhu, Yu Qiao, Ziwei Liu~\textsuperscript{\Letter}
\IEEEcompsocitemizethanks{
\IEEEcompsocthanksitem \textsuperscript{\Letter} Corresponding author.
\IEEEcompsocthanksitem Email: \textnormal{\texttt{\{shulin002, ziqi.huang, ziwei.liu\}@ntu.edu.sg}}
\IEEEcompsocthanksitem A preliminary version of this work was published in ACL 2025~\cite{zhang2025evaluationagent}.
\IEEEcompsocthanksitem The authors are with S-Lab, Nanyang Technological University; Agency for Science, Technology and Research (A*STAR); Shanghai Artificial Intelligence Laboratory.
}
}

\markboth{Preprint}%
{Tian \MakeLowercase{\textit{et al.}}: Open Evaluation Agent for Visual Generative Models}
\IEEEtitleabstractindextext{%
\begin{abstract}
Recent advancements in visual generative models have enabled high-quality image and video generation, opening diverse applications. However, evaluating these models often demands sampling hundreds or thousands of images or videos, making the process computationally expensive, especially for diffusion-based models with inherently slow sampling. Moreover, existing evaluation methods rely on rigid pipelines that overlook specific user needs and provide numerical results without clear explanations. In contrast, humans can quickly form impressions of a model's capabilities by observing only a few samples. To mimic this, we propose the \textbf{Evaluation Agent} framework, which employs human-like strategies for efficient, dynamic, multi-round evaluations using only a few samples per round, while offering detailed, user-tailored analyses. Given a natural-language evaluation request, the agent decomposes it into sub-aspects, generates targeted prompts, samples images or videos from the evaluated model, invokes suitable evaluation tools, and iteratively updates its plan from the observed evidence. This tool-grounded loop enables the framework to evaluate both predefined benchmark dimensions and open-ended user concerns without requiring a fixed prompt set or a single handcrafted metric. It offers four key advantages: \textit{1)} \textbf{efficiency}, \textit{2)} \textbf{promptable evaluation} tailored to diverse user needs, \textit{3)} \textbf{explainability} beyond single numerical scores, and \textit{4)} \textbf{scalability} across various models and tools. Experiments show that Evaluation Agent reduces evaluation time to 10\% of traditional methods while delivering comparable results. We introduce \textbf{Open Evaluation Agent (Open-EA)} by constructing EA-CoT-10K, a corpus of history-conditioned step-level instruction-tuning records derived from multi-round evaluation rollouts, and training EA-3B from Qwen2.5-3B-Instruct~\cite{yang2024qwen2technicalreport} as a local planning backbone. The resulting model preserves the structured reasoning, tool invocation, observation analysis, and summary protocol of the API-based agent, while reducing dependence on proprietary planning backbones. Experiments validate the API-based Evaluation Agent on established T2I/T2V benchmarks and open-ended queries, and evaluate Open-EA on four in-domain and three out-of-domain T2V generator families. The results demonstrate efficient, promptable, and interpretable visual-model evaluation while showing partial cross-family transfer of the learned Open-EA policy. We release the code, data, and models through our GitHub repository at \href{https://github.com/Vchitect/Evaluation-Agent}{\nolinkurl{github.com/Vchitect/Evaluation-Agent}}.

\end{abstract}
}

\maketitle

\begin{table*}[!t]
\centering
\caption{\textbf{Comparison of Open-EA with Traditional T2I and T2V Benchmarks.} Open-EA supports customized user queries in natural language and works with both T2I and T2V models. Unlike traditional benchmarks, it dynamically updates the evaluation process using multiple tools, providing comprehensive and explainable results with detailed textual analysis. The Open-EA row encompasses both the API-based Evaluation Agent and the local EA-3B instantiation. The released EA-3B checkpoint is trained and evaluated on T2V, while T2I records form a separate companion set.}
\label{tab:related_work_comp}
\resizebox{\textwidth}{!}{%
\begin{tabular}{c|c|c|c|c|c|c|c}
\toprule
\textbf{Benchmark} & \textbf{\makecell{Analysis}} & \textbf{\Centerstack{Customized \\Queries}} & \textbf{\Centerstack{Supported \\Models}} & \textbf{\Centerstack{\# Required\\Samples}} & \textbf{\Centerstack{Open Evaluation\\Request Support}} & \textbf{\Centerstack{Dynamic \\Evaluation}} & \textbf{\Centerstack{Open \\Tool-Use}} \\
\midrule
FID / FVD~\cite{unterthiner2018towards, heusel2017gans} & \xmark & \xmark & T2I / T2V & 2,048 & \xmark~(Fixed-Form) & \xmark & \xmark\\
T2I-CompBench~\cite{huang2023t2icompbench} & \xmark & \xmark & T2I & 18,000 & \xmark~(Pre-Defined) & \xmark & \xmark\\
VBench~\cite{huang2023vbench} & \xmark & \xmark & T2V & 4,730 & \xmark~(Pre-Defined) & \xmark & \xmark\\
\midrule
\textbf{Open-EA (Ours)} & \cmark & \cmark & T2I \& T2V & $\approx 400$ & \cmark~(Open-Ended) & \cmark & \cmark\\
\bottomrule
\end{tabular}%
}
\vspace{-15pt}
\end{table*}

\section{Introduction}
\label{intro}

\IEEEPARstart{V}{isual} generative models have made significant progress in recent years, particularly driven by the advancement of diffusion models~\cite{ho2020denoising} and the availability of internet-scale datasets~\cite{Bain21, chen2024panda70m, xue2019video}. These advancements enable the generation of high-quality images and videos, opening up a wide range of applications in content creation, design inspiration, and beyond.

With the advancement of visual generative models, effective evaluation is crucial for understanding their strengths, limitations, and areas for improvement. Existing evaluation frameworks, such as VBench~\cite{huang2023vbench, huang2024vbench++}, EvalCrafter~\cite{liu2023evalcrafter}, and T2I-CompBench~\cite{huang2023t2icompbench}, assess models across multiple dimensions using specific prompts and tailored metrics to ensure comprehensive performance analysis. However, these approaches often demand generating numerous samples, resulting in long evaluation time and high computational costs, particularly for diffusion-based models where sampling is inherently slow due to iterative sampling. Furthermore, these evaluation frameworks are constrained by rigid evaluation pipelines and predefined dimensions, making them less adaptable to open-ended inputs or diverse user needs. Additionally, these methods often produce single numerical scores as outcomes, requiring users to invest additional effort to extract meaningful insights.

In contrast, human evaluators can quickly gain a general understanding of a model's performance by interactively testing a few prompts, forming a sufficient impression without taking too much time. This type of evaluation has several unique advantages for assessing visual generative models. First, it is \textbf{fast}, requiring only a small number of samples to assess overall performance. Second, it is \textbf{flexible}, allowing intuitive evaluation of various aspects, such as realism, creativity, prompt adherence, or other user-defined criteria. Third, it is \textbf{dynamic}, enabling deeper, hierarchical evaluations through continuous adjustments in exploration.

To leverage the strengths of human-like evaluations, we introduce the \textbf{Evaluation Agent}~\cite{zhang2025evaluationagent}, a paradigm that mimics human strategies for assessing visual generative models. The Evaluation Agent offers four key features: \textbf{\textit{1) Efficiency}:} It dynamically adjusts its evaluation pathway based on intermediate results, uncovering subtle model behaviors and limitations while avoiding redundant test cases for efficient evaluation. \textbf{\textit{2) Promptable Evaluation}:} Unlike existing benchmarks with fixed prompts and evaluation metrics, it accepts open-ended user input, allowing for flexible and customized assessments tailored to specific user needs. \textbf{\textit{3) Detailed and Interpretable Results}:} It provides interpretable, detailed insights beyond single numerical scores, making results accessible to both experts and non-experts. \textbf{\textit{4) Scalability}:} The framework supports seamless integration of new metrics and evaluation tools, ensuring adaptability and growth.

The Evaluation Agent begins by accepting open-ended user input, specifying what to evaluate and which model(s) to assess. Based on this input, it identifies initial evaluation aspects and leverages appropriate tools to conduct the assessment. It then observes the intermediate results and dynamically refines the direction of further exploration. In the end, it generates a detailed natural language response summarizing the evaluation results, providing a comprehensive analysis of the evaluation process and a clear summary of the model's capabilities as specified in the user input.
The Evaluation Agent can also automate various applications, including: \textit{1) Model Comparison}: Allowing users to compare models based on specific criteria to determine which performs better in a given aspect. \textit{2) Model Recommendation}: Suggesting the most suitable model for the user's needs by evaluating models against personalized criteria.

We demonstrate the versatility of the Evaluation Agent through experiments on diverse scenarios, including the evaluation of image and video generation models. The results indicate that it delivers performance comparable to full benchmark pipelines while significantly reducing evaluation time. Furthermore, we create an open-ended user query dataset to showcase the Evaluation Agent's flexibility, depth, and accuracy in addressing open-ended queries.

To reduce dependence on API-based agents, we study whether the evaluation behavior can be transferred to a compact open-source backbone. We build EA-CoT-10K by unfolding recorded multi-round evaluation rollouts into history-conditioned, step-level instruction-tuning records that preserve the user query, intermediate observations, reasoning traces, tool-use decisions, and final summaries. We then fine-tune Qwen2.5-3B-Instruct into EA-3B using LLaMA-Factory~\cite{zheng2024llamafactory}, allowing the same evaluation protocol to be executed by a local model with structured reasoning and tool-call outputs. We refer to this open-source local instantiation as \textbf{Open Evaluation Agent (Open-EA)}.

We summarize our contributions as follows:
\begin{itemize}[noitemsep, topsep=0pt]
    \item We propose the \textbf{Evaluation Agent}, a human-like evaluation framework that overcomes the limitations of existing methods in flexibility and efficiency. Our approach will be fully open-sourced.

    \item We release \textbf{EA-CoT-10K}, a first-of-its-kind dataset comprising 10,042 history-conditioned T2V step-level instruction-tuning records derived from multi-round evaluation rollouts, together with a separate 986-record T2I companion set. The corpus captures the decision-making logic of large-scale proprietary models, serving as a critical resource for training autonomous evaluation systems.

    \item We develop \textbf{EA-3B}, a compact open-source planning backbone for Open-EA. Across four in-domain and three out-of-domain T2V generator families, EA-3B follows the same reasoning, tool-selection, observation, and summary protocol as Evaluation Agent and exhibits partial cross-family transfer while reducing dependence on external model APIs.

    \item We validated our approach on several widely adopted benchmarks, demonstrating that it achieves evaluation accuracy comparable to standard benchmarks while reducing evaluation time by over 90\%. We also built an open-ended user query dataset to demonstrate our method's flexibility, depth, and accuracy in handling open-ended evaluation queries.
    \item Through a comprehensive analysis of how standard benchmarks, human evaluators, and our Evaluation Agent perform evaluations, we provide in-depth insights that serve as important cornerstones for future research in evaluating visual generative models.
\end{itemize}

\section{Related Work}
\label{sec:related}

\subsection{Visual Generation and Evaluation}

Visual generative models~\cite{rombach2022high, podell2023sdxl, esser2024scaling, ma2024latte, wang2023modelscope, chen2024videocrafter2, he2022lvdm, wang2025lavie} have gained significant attention in recent years. However, unlike perception tasks, which have clear evaluation metrics such as accuracy, evaluating visual generative tasks is more challenging due to the absence of a definitive ``ground truth'' or single correct answer. Metrics such as FID~\cite{heusel2017gans} and FVD~\cite{unterthiner2018towards} are commonly used to measure the distance between generated samples and reference datasets. Recent benchmarks~\cite{huang2023t2icompbench, huang2023vbench, zheng2025vbench2, huang2024vbench++, liu2023evalcrafter, lee2023holistic, sun2024t2v} provide multi-dimensional evaluations tailored to specific model capabilities, while benchmarks for unified multimodal models probe the coupling between understanding and generation~\cite{shi2026realunify, zou2026unimmmu, liu2026unison}. However, whether relying on metrics like FID~\cite{heusel2017gans} and FVD~\cite{unterthiner2018towards} or predefined benchmarks such as VBench, these protocols use fixed test sets and scoring pipelines rather than adaptively choosing what to probe next.

\subsection{LLM as a Judge}
Recent advances in the understanding and reasoning capabilities of Large Language Models (LLMs) have enabled their use as powerful evaluators~\cite{Jain_2023, chiang2023largelanguagemodelsalternative, fu2023gptscoreevaluatedesire, zheng2023judging}. For instance, Li et al.~\cite{li2023collaborativeevaluationexploringsynergy} study customized LLM evaluators across diverse NLP tasks and show that LLM-generated criteria can reduce annotation variance while still requiring careful human scrutiny. Pan et al.~\cite{pan2024autonomous} demonstrate how LLM-based evaluations enhance downstream tasks for digital agents. These studies showcase the ability of LLMs to reason about criteria and explain judgments. However, they typically score already supplied responses or trajectories under a fixed judging prompt, rather than deciding what evidence to acquire or which evaluation action to execute next. Moreover, devising effective strategies for domain-specific problems remains challenging~\cite{wang2024survey}. A recent survey systematizes this literature by what to judge, how to judge, and how to benchmark, and identifies dynamic examiner-style pipelines and complex judging agents as an emerging direction~\cite{li2025generationjudgment}.

\subsection{Agentic Evaluation}
Beyond single-pass LLM judges, recent work frames the evaluator itself as an agent that decomposes objectives, acquires evidence, and inspects intermediate states. A recent survey organizes this shift into procedural, reactive, and self-evolving judge agents, spanning multi-agent collaboration, planning, tool integration, memory, and optimization~\cite{you2026agentasjudge}. Zhuge et al.'s Agent-as-a-Judge evaluates complete code-agent trajectories against hierarchical requirements~\cite{zhuge2025agentasjudge}. Task elicitation adaptively proposes and verifies questions to discover target-model failure modes~\cite{brown2025taskelicitation}, while Mind2Web~2 instantiates task-specific judge agents from tree-structured rubrics for long-horizon, time-varying search tasks~\cite{gou2025mind2web2}. One-Eval instead compiles natural-language evaluation requests into traceable benchmark, metric, execution, and reporting workflows~\cite{shen2026oneeval}. For safety evaluation, AgenticEval turns regulatory documents and evaluation feedback into a self-evolving LLM-safety test suite~\cite{wang2026agenticeval}.

Agentic evaluation has also expanded to visual generation. Scene-graph question answering diagnoses hallucinations in text-to-image models~\cite{qin2024evaluating}, and CIGEval decomposes conditional-image evaluation into fine-grained checks and selects specialized visual tools~\cite{wang2025cigeval}. For image editing, EdiVal-Agent combines object-centric decomposition with specialist evaluators for multi-turn evaluation~\cite{chen2026edival}, while RewardHarness evolves a library of tools and skills from preference demonstrations to judge instruction-guided edits~\cite{zhang2026rewardharness}. For video, VideoGen-Eval combines LLM-based content structuring, MLLM judgment, and temporal patch tools in a dynamic evaluation pipeline~\cite{yang2025videogen}; VQQA generates visual questions and uses the resulting critiques to refine generation prompts~\cite{song2026vqqa}. Most recently, VideoArgus constructs output-blind, input-specific rubrics and executes criterion-specific evidence plans across video generation and editing settings~\cite{zeng2026videoargus}. These systems primarily operate at the instance or predefined-task level, with some additionally optimizing a particular generated sample. In contrast, Evaluation Agent treats the generator itself as the object of investigation: from an open-ended user request, it repeatedly designs prompts, samples new outputs, selects heterogeneous tools, and changes the next probe or termination decision according to accumulated observations.

The reliability of automated evaluators has consequently become a first-class evaluation target. AgentRewardBench finds that no single LLM judge is consistently strongest across web-agent benchmarks~\cite{lu2025agentrewardbench}, while AJ-Bench directly evaluates judge agents' ability to acquire information and verify environment states and execution processes~\cite{shi2026ajbench}. Controlled meta-evaluation in REFLECT further exposes substantial difficulty in detecting reasoning, tool-use, and evidence failures in research-agent trajectories~\cite{wang2026reflect}. More broadly, audits of agentic benchmarks show that task setup, reward design, and outcome-verification procedures can materially distort measured performance~\cite{zhu2025rigorousagentbenchmarks}. These findings motivate tool-grounded, auditable evaluation traces and caution against treating model-based judgments as infallible ground truth.

\subsection{Agent in Planning \& Reasoning}

Agents and agentic systems are gaining attention for their ability to automate complex tasks and design customized trajectories based on user queries. They have been explored across various domains, including web, mobile, desktop, and operating systems (OS) ~\cite{zhou2023webarena, xie2024osworld, wang2024mobile, kapoor2024omniact, tian2025mmina}, showing effectiveness in improving long-horizon task completion. For example, Chain-of-Thought (CoT)~\cite{wei2022chain} and Zero-shot-CoT~\cite{kojima2022large} use prompting techniques to enable step-by-step reasoning. Similarly, ReAct~\cite{yao2022react} introduces a general paradigm for agent prompting by integrating reasoning traces with task-specific actions through interleaved triplets of ``thought-action-observation,'' thereby incorporating environmental feedback. This action-observation loop provides a natural foundation for an evaluator that revises its tests in response to intermediate evidence.
Beyond single-chain reasoning, recent methods explore richer deliberation structures, including Tree of Thoughts~\cite{yao2024tree}, Algorithm of Thoughts~\cite{sel2023algorithm}, Graph of Thoughts~\cite{besta2024graph}, self-consistency~\cite{wang2022self}, and Reflexion~\cite{shinn2024reflexion}. Tool-use oriented systems and benchmarks, such as API-Bank~\cite{li2023api}, ToolLLM~\cite{qin2023toolllm}, Toolformer~\cite{schick2024toolformer}, and Gorilla~\cite{patil2023gorillalargelanguagemodel}, further show that explicit action selection is important for agents operating over external resources. Beyond evaluation, similar reasoning paradigms have also been used to improve generation itself~\cite{jiang2026t2i, huang2026vchain}. Evaluation Agent adopts this reasoning-and-action view but grounds each action in visual generation sampling or benchmark/VQA-based assessment.

Recent agent benchmarks have likewise shifted toward stateful, executable, and auditable evaluation. $\tau^2$-Bench models both the agent and the user as tool-using actors that modify a shared environment~\cite{barres2025tau2bench}; Online-Mind2Web evaluates agents on live websites and shows that offline benchmarks can overstate capability~\cite{xue2025illusion}; and PaperBench combines hierarchical rubrics co-developed with paper authors with a judge evaluated on a separate benchmark for long-horizon research replication~\cite{starace2025paperbench}. These benchmarks primarily evaluate agents as task performers, whereas our framework uses an agent to adaptively design and execute evaluations of visual generative models.

\section{Methods}
\label{methods}

\begin{figure*}[t]
\centering
\includegraphics[width=\textwidth]{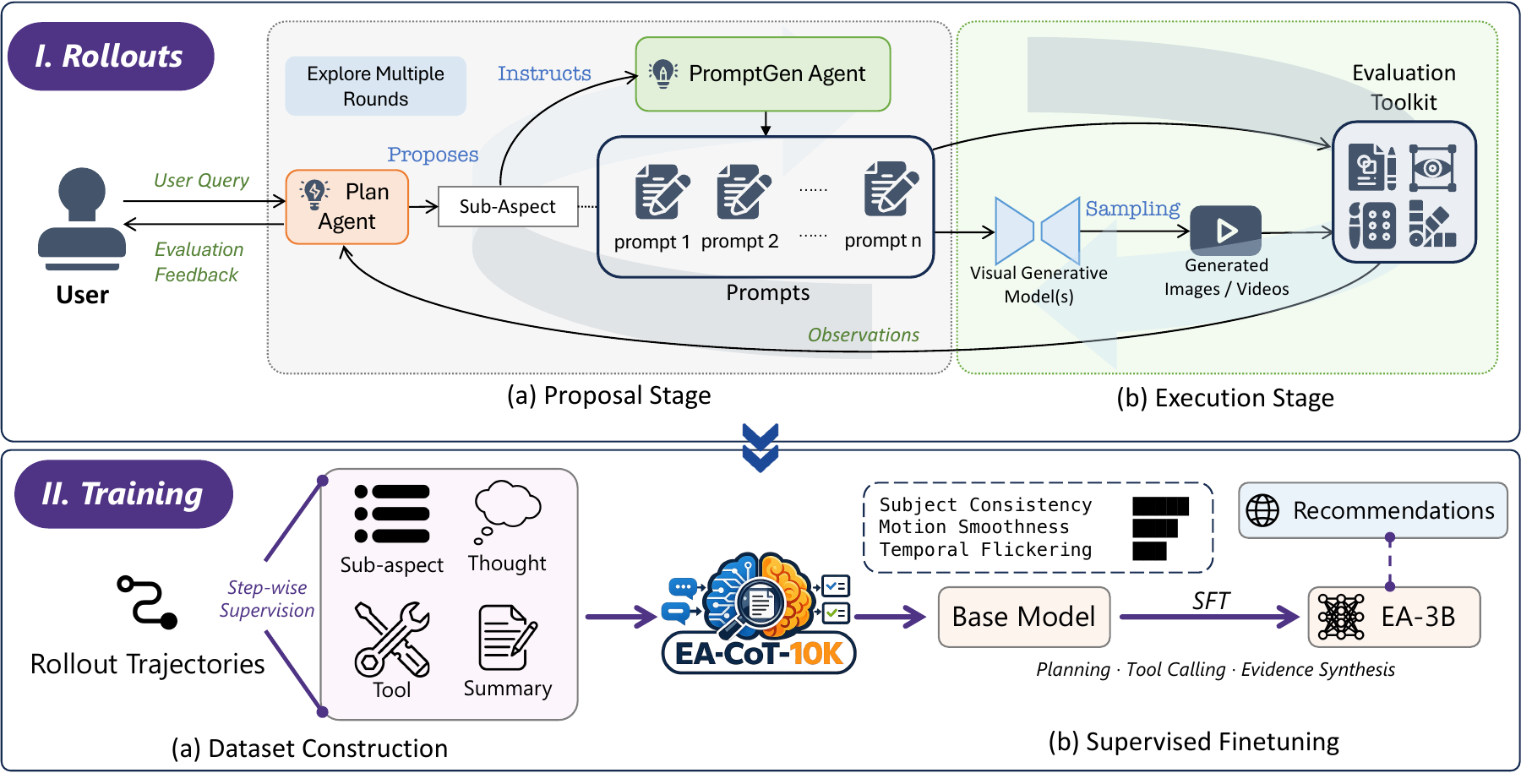}
\caption{\textbf{Overview of the Evaluation Agent Framework.} The upper rollout pipeline contains a proposal stage, where the Plan Agent decomposes the user request and the PromptGen Agent prepares evaluation prompts, and an execution stage, where visual outputs are sampled and assessed by selected tools. These stages form an iterative observation loop for dynamic, user-conditioned evaluation. The bottom training pipeline converts completed rollouts into step-wise supervision. Sub-aspects, thoughts, tool calls, observations, and final summaries are organized into EA-CoT-10K and used to fine-tune a base model into EA-3B, allowing Open-EA to follow the same evaluation protocol with a compact local planning backbone.}
\label{fig:pipeline}
\end{figure*}

\subsection{Preliminaries: Evaluation of Visual Generative Models}

\noindent\textbf{Evaluation Benchmark.} Let $C=\{c_{j} \mid j \in \{1, 2, 3, \ldots, N\}\}$ denote the condition set (\ie, test cases), whose elements can be text prompts, input images, class labels, or conditions in other formats. For unconditional generation, $C$ is empty or consists of random seeds. Existing evaluation approaches typically pre-define hundreds or thousands of conditions, making visual sampling computationally expensive. In the Evaluation Agent framework, $C$ is determined dynamically during evaluation and usually contains only a small number of cases.

\noindent\textbf{Sampling.} $v_{j} = G(c_{j})$
where $G$ is the visual generative model, which generates the visual output $v_j$ (\ie, images or videos) given an optional condition $c_j$. $V = G(C)$
where $V=\{v_j \mid j \in \{1, 2, 3, \ldots, N\}\}$ represents the set of generated visuals for the entire condition set $C$.

\noindent\textbf{Evaluation Pipeline.} Existing evaluation methods usually follow a fixed pipeline to evaluate all the images or videos $V$ sampled from the pre-defined benchmark $C$.
\begin{equation}
    y_j = e_k(v_j, c_j)
\end{equation}
where $e_k \in E$ is an evaluation function for an aspect such as aesthetics or compositionality. Given the generated visual $v_j$ and the optional condition $c_j$, it produces the evaluation result $y_j$.

Some reference/statistics-based evaluation frameworks like FID and FVD also use reference datasets $V_r$ for calculating the results.
\begin{equation}
    Y = E(V, V_r, C)
\end{equation}
In existing evaluation approaches, $E$ is a pre-defined set of evaluation dimensions, or a single evaluation metric, which limits the possible evaluation aspects from the beginning, and requires assessing all the aspects even if some are not needed in some cases. In our Evaluation Agent framework, the evaluation tool $e_k$ is dynamically determined during the evaluation process.

\begin{table*}[!t]
\caption{\textbf{Evaluation Results Comparison with VBench~\cite{huang2023vbench}}. We evaluated 15 specific ability dimensions in VBench using our Evaluation Agent and compared its results against VBench in terms of conclusion accuracy. The numerical results show the percentages of the Evaluation Agent's correct predictions falling either within the exact range (left) or within an error margin of one range (right) across ten trials.}
    \centering
    \small
    \setlength{\tabcolsep}{2pt}
    \resizebox{\textwidth}{!}{%
        \begin{tabular}{>{\centering\arraybackslash}p{4.5cm}|
                            >{\centering\arraybackslash}p{2cm}|
                            >{\centering\arraybackslash}p{2cm}|
                            >{\centering\arraybackslash}p{2cm}|
                            >{\centering\arraybackslash}p{2cm}|
                            >{\centering\arraybackslash}p{2cm}|
                            >{\centering\arraybackslash}p{2cm}|
                            >{\centering\arraybackslash}p{2cm}}
        \Xhline{1pt}
        \textbf{Models} & \textbf{\Centerstack{Subject\\Consistency}} & \textbf{\Centerstack{Background\\Consistency}} &
        \textbf{\Centerstack{Motion\\Smoothness}} &
        \textbf{\Centerstack{Dynamic\\Degree}} &
        \textbf{\Centerstack{Aesthetic\\Quality}} &
        \textbf{\Centerstack{Imaging\\Quality}} &
        \textbf{\Centerstack{Object\\Class}} \\
        \Xhline{1pt}
        Latte-1~\cite{ma2024latte} & 50{\%} / 80{\%} & 0{\%} / 30{\%} & 40{\%} / 70{\%} & 30{\%} / 70{\%} & 60{\%} / 100{\%} & 70{\%} / 100{\%} & 40{\%} / 50{\%} \\
        ModelScope~\cite{wang2023modelscope}   & 80{\%} / 80{\%} & 80{\%} / 90{\%} & 60{\%} / 80{\%} & 60{\%} / 100{\%} & 60{\%} / 100{\%} & 100{\%} / 100{\%} & 0{\%} / 50{\%} \\
        VideoCrafter-0.9~\cite{he2022lvdm}  & 100{\%} / 100{\%} & 80{\%} / 100{\%} & 70{\%} / 100{\%} & 80{\%} / 100{\%} & 90{\%} / 100{\%} & 20{\%} / 100{\%} & 20{\%} / 60{\%} \\
        VideoCrafter-2~\cite{chen2024videocrafter2}  & 10{\%} / 100{\%} & 60{\%} / 100{\%} & 30{\%} / 90{\%} & 30{\%} / 80{\%} & 80{\%} / 100{\%} & 50{\%} / 100{\%} & 70{\%} / 100{\%} \\
        \Xhline{1pt}
        \end{tabular}
    }

    \vspace{5pt}

    \setlength{\tabcolsep}{2pt}
    \resizebox{\textwidth}{!}{%
    \begin{tabular}{>{\centering\arraybackslash}p{4.5cm}|
                            >{\centering\arraybackslash}p{2cm}|
                            >{\centering\arraybackslash}p{2cm}|
                            >{\centering\arraybackslash}p{2cm}|
                            >{\centering\arraybackslash}p{2cm}|
                            >{\centering\arraybackslash}p{2cm}|
                            >{\centering\arraybackslash}p{2cm}|
                            >{\centering\arraybackslash}p{2cm}}
    \Xhline{1pt}
    \textbf{\Centerstack{Multiple\\Objects}} &
    \textbf{\Centerstack{Human\\Action}} &
    \textbf{\Centerstack{Color}} &
    \textbf{\Centerstack{Spatial\\Relationship}} &
    \textbf{\Centerstack{Scene}} &
    \textbf{\Centerstack{Appearance\\Style}} &
    \textbf{\Centerstack{Temporal\\Style}} &
    \textbf{\Centerstack{Overall\\Consistency}} \\
    \Xhline{1pt}
    40{\%} / 100{\%} & 10{\%} / 10{\%} & 30{\%} / 70{\%} & 10{\%} / 80{\%} & 20{\%} / 40{\%} & 70{\%} / 90{\%} & 40{\%} / 100{\%} & 70{\%} / 100{\%} \\
    50{\%} / 100{\%} & 10{\%} / 40{\%} & 0{\%} / 20{\%} & 10{\%} / 30{\%} & 20{\%} / 100{\%} & 90{\%} / 100{\%} & 50{\%} / 90{\%} & 20{\%} / 100{\%} \\

    80{\%} / 100{\%} & 10{\%} / 30{\%} & 10{\%} / 40{\%} & 20{\%} / 100{\%} & 30{\%} / 100{\%} & 60{\%} / 100{\%} & 80{\%} / 100{\%} & 0{\%} / 80{\%} \\
    20{\%} / 60{\%} & 10{\%} / 90{\%} & 90{\%} / 100{\%} & 0{\%} / 70{\%} & 0{\%} / 10{\%} & 80{\%} / 100{\%} & 80{\%} / 100{\%} & 60{\%} / 100{\%} \\
    \Xhline{1pt}
    \end{tabular}
    }
    \label{tab:main_results_t2v}
    \vspace{-5pt}
\end{table*}

\begin{table*}[htbp]
    \caption{\textbf{Evaluation Results Comparison with T2I-CompBench~\cite{huang2023t2icompbench}}. We evaluated four ability dimensions in T2I-CompBench using our Evaluation Agent and compared its results with those of T2I-CompBench in terms of conclusion accuracy. The numerical results show the percentages of the Evaluation Agent's correct predictions falling either within the exact range (left) or within an error margin of one range (right) across ten trials.}    \centering
    \small
    \setlength{\tabcolsep}{3pt}
    \begin{tabular}{c|c|c|c|c}
    \Xhline{1pt}
    \textbf{Models} & \textbf{\Centerstack{Color\\Binding}} &     \textbf{\Centerstack{Shape\\Binding}} &
    \textbf{\Centerstack{Texture\\Binding}} &     \textbf{\Centerstack{Non-Spatial\\Relationships}} \\ \Xhline{1pt}
    SD1.4~\cite{rombach2022high}        & 50{\%} / 100{\%} & 100{\%} / 100{\%} & 0{\%} / 100{\%} & 50{\%} / 100{\%} \\
    SD2.1 ~\cite{rombach2022high}  & 100{\%} / 100{\%} & 60{\%} / 100{\%} & 80{\%} / 100{\%} & 60{\%} / 100{\%} \\
    SDXL ~\cite{podell2023sdxl}  & 100{\%} / 100{\%} & 20{\%} / 100{\%} & 80{\%} / 100{\%} & 60{\%} / 100{\%} \\
    SD3.0 ~\cite{esser2024scaling}  & 20{\%} / 90{\%} & 0{\%} / 90{\%} & 0{\%} / 70{\%} & 80{\%} / 90{\%} \\
    \Xhline{1pt}
    \end{tabular}
    \vspace{-10pt}
    \label{tab:main_results_t2i}
\end{table*}

\subsection{The Evaluation Agent Framework}

Our Evaluation Agent framework is powered by LLM-based agents, leveraging their advanced planning capabilities to simulate human-like behaviors for efficient and flexible visual model assessments. As illustrated in Figure~\ref{fig:pipeline}, the rollout component of the framework operates in two stages: the proposal stage and the execution stage. By iteratively interacting and looping between these stages, the framework dynamically evaluates models based on user queries. A training component reuses the rollout traces as supervision. Each trajectory records the selected sub-aspects, reasoning thoughts, tool interactions, observations, and summaries; these structured records are assembled into EA-CoT-10K and used to train EA-3B. This connects the API-based Evaluation Agent with Open-EA, where the trained local backbone follows the same propose--sample--evaluate--observe protocol during deployment.

\subsubsection{Proposal Stage}
\label{sssec:proposal_stage}

The Proposal Stage consists of two agents: the Plan Agent and the PromptGen Agent. The Plan Agent is responsible for planning, observing, and summarizing the evaluation process based on the user's query, while the PromptGen Agent focuses specifically on the design aspects.

\noindent\textbf{Plan Agent.} We design the Plan Agent to simulate human behavior during the evaluation process, including planning and adjusting the evaluation direction, observing intermediate results, and summarizing the final outcomes. As the core component of the framework, the Plan Agent not only interacts with the user but also drives the entire evaluation process. Specifically, upon receiving a user query, the Plan Agent identifies an initial sub-aspect for evaluation and iteratively refines it based on feedback from intermediate results. This process continues until sufficient information is collected, after which the agent provides a detailed analysis and summary.
At each step, the Plan Agent is required to provide an explicit rationale for the selected sub-aspect and, when it terminates the loop, to justify why the collected observations are sufficient to answer the user query. For closed-domain benchmark queries, it also selects the corresponding evaluation tool; for open-ended queries, it formulates the evaluation direction that will later be converted into VQA-style observations.

\noindent\textbf{PromptGen Agent.} The PromptGen Agent mimics human behavior in designing prompts for visual generative models based on the plan developed during the evaluation process. Specifically, it generates tailored prompts for each sub-aspect identified by the Plan Agent, enabling focused content generation and evaluation. Additionally, the PromptGen Agent can reference and utilize well-established prompts from existing benchmarks.
In benchmark-based evaluation, PromptGen selects prompts from the official prompt lists and keeps the evaluation comparable to the original benchmark protocol. In open-ended evaluation, it jointly designs generation prompts and VQA questions so that each sampled visual output can be converted into textual evidence for the Plan Agent.

\subsubsection{Execution Stage}
\label{sssec:execution_stage}

The Execution Stage is responsible for sampling and evaluating the model using the appropriate tools, as specified in the Proposal Stage, and for returning the final evaluation results.

\noindent\textbf{Visual Generative Models.} This component takes prompts designed by the PromptGen Agent as input and generates corresponding visual content, which is then used for subsequent evaluation.

\noindent\textbf{Evaluation Toolkit.} The Evaluation Toolkit consists of a set of elementary evaluation tools for visual generative models. This module is open and extensible, allowing for continuous expansion. We have integrated several existing evaluation tools from well-known benchmarks for different modalities of visual generation models. To support the evaluation of open-ended user queries, we have introduced a paradigm based on vision-language models (VLMs), leveraging the Visual Question Answering (VQA) format to enable flexible assessments across various aspects of the models. Upon receiving the generated samples from visual generative models along with the corresponding prompts, the module utilizes the appropriate tools to evaluate each sample. All evaluation results are then compiled and returned to the Plan Agent for further proposals or summarization.
The implemented toolkit covers VBench dimensions for text-to-video evaluation, T2I-CompBench tools for compositional text-to-image evaluation, and a VLM-based VQA tool for open-ended image-model evaluation. This design keeps the tool interface modular: each evaluation round returns structured observations rather than only final benchmark scores.

\subsubsection{Overall Pipeline}
\label{sssec:overall_pipeline}

The Evaluation Agent's process is dynamic and multi-round, with each round comprising a proposal stage and an execution stage. By interacting and looping through these stages, we achieve dynamic evaluation, where the evaluation process adapts based on intermediate observations and initial user query. This dynamic approach allows the Evaluation Agent to refine its focus iteratively, adjusting its exploration direction and prompt design based on an evolving understanding of the model's capabilities. Consequently, the evaluation process becomes more efficient and targeted, systematically identifying the strengths and limitations of generative models.

Operationally, each round follows a propose--sample--evaluate--observe loop. The agent proposes one sub-aspect, samples a small set of images or videos, invokes one selected tool, receives per-sample or aggregate observations, and then decides whether to continue probing or produce a final answer. This loop allows the evaluation to expand breadth-wise across categories or depth-wise toward harder cases, depending on the evidence gathered in previous rounds.

\subsection{EA-CoT Data Construction and Open-EA Training}
\label{ssec:ea_cot_training}

\begin{figure*}[t]
    \centering
    \includegraphics[width=\textwidth]{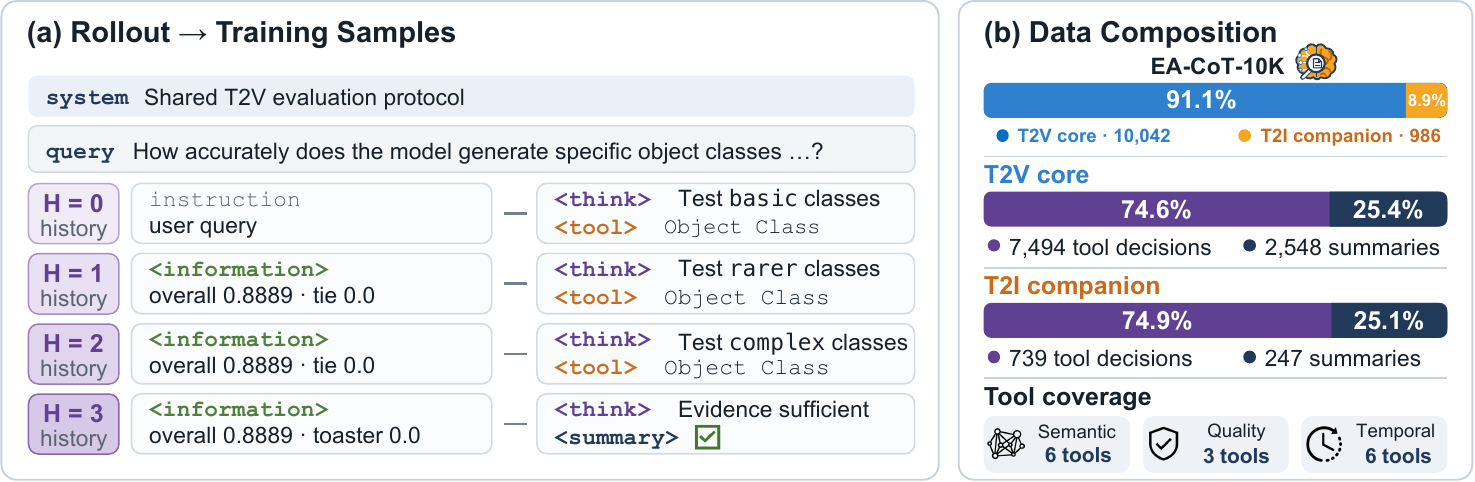}
    \caption{\textbf{EA-CoT-10K Construction and Data Composition.} (a) A representative T2V evaluation rollout is unfolded into history-conditioned step-level instruction-tuning records (training samples). Under a shared system protocol, each record retains preceding turns in its history and supervises either the next reasoning-and-tool decision or, after sufficient evidence has been collected, the terminal summary; text is abbreviated for clarity. (b) The upper bar compares the relative sizes of the separately processed T2V core (10,042 records; 91.1\%) and T2I companion set (986 records; 8.9\%). The lower bars show their within-set tool-decision/summary composition: 7,494/2,548 (74.6\%/25.4\%) for T2V and 739/247 (74.9\%/25.1\%) for T2I. The T2V core covers 15 VBench tools grouped into six semantic, three quality, and six temporal tools.}
    \label{fig:eacot_structure}
\end{figure*}

\noindent\textbf{EA-CoT Data Construction.} To transfer the evaluation protocol into a trainable local agent, we construct EA-CoT from recorded evaluation trajectories rather than isolated question-answer pairs. Each trajectory preserves the user query, the previous observations available to the agent, the reasoning segment used to select the next sub-aspect, the selected tool when applicable, the returned information, and the final evidence-based summary. The post-processing pipeline converts these trajectories into instruction-tuning records with explicit tags such as \texttt{<think>}, \texttt{<tool>}, \texttt{<information>}, and \texttt{<summary>}. This format exposes both the intermediate decision process and the terminal evaluation judgment to supervised training.
As illustrated in Figure~\ref{fig:eacot_structure}(a), post-processing unfolds each rollout into history-conditioned step-level instruction-tuning records, each serving as a training sample whose target is either the next reasoning-and-tool decision or the terminal summary. Figure~\ref{fig:eacot_structure}(b) compares the scale and within-set record composition of the T2V core and the separate T2I companion set and summarizes the T2V tool coverage.

\noindent\textbf{Data Filtering.} Before packaging EA-CoT-10K, we apply a two-stage filtering procedure to improve the faithfulness of the supervision. First, we perform deterministic score-based filtering using the tool scores stored in each rollout and the corresponding dimension-specific scoring reference. Trajectories whose generated images or videos exhibit severe prompt-visual misalignment, such as extremely low aggregate scores or per-prompt scores in the lowest-quality range for the selected dimension, are removed because their visual evidence is too noisy to support reliable reasoning. Second, we deploy an MLLM as a judge to verify whether the textual reasoning is grounded in the original visual outputs. For each candidate trajectory, the judge is given the user query, generated prompts, sampled visual evidence, tool observations, and the agent's reasoning and summary. It then checks whether the stated objects, actions, attributes, temporal changes, and final conclusion are supported by the generated video or image. We discard trajectories where the reasoning contradicts the visual content, hallucinates absent details, overlooks obvious generation failures, or maps the score evidence to an inconsistent qualitative judgment.
In our processed artifacts, the T2V table-context variant contains 10,042 history-conditioned step-level records (7,494 tool-decision records and 2,548 summary records) spanning 15 VBench evaluation tools. A separate T2I companion set contains 986 records (739 tool-decision records and 247 summary records) across four T2I-CompBench tools. The companion set is not mixed into the current EA-3B training run. Each Alpaca-style record contains \texttt{instruction}, \texttt{input}, \texttt{output}, \texttt{system}, and \texttt{history} fields.

\noindent\textbf{Open-EA Training.} We obtain EA-3B by full-parameter supervised fine-tuning of Qwen2.5-3B-Instruct~\cite{yang2024qwen2technicalreport} on the 10,042-record T2V core using LLaMA-Factory~\cite{zheng2024llamafactory}. We train for three epochs with a maximum sequence length of 4096, an effective batch size of 64, and a learning rate of $1\times10^{-5}$, using cosine decay and a warmup ratio of 0.1. The structured reasoning, tool, information, and summary fields teach EA-3B to choose the next evaluation action from accumulated evidence and to terminate with a grounded assessment. EA-3B thereby provides Open-EA with a compact local planning backbone while preserving the interaction protocol and tool workflow of the API-based Evaluation Agent.

\section{Experiments}
\label{sec:experiments}

We first validate the efficiency of our Evaluation Agent on established benchmarks for visual generative models and then demonstrate the flexibility, depth, and accuracy of our approach in handling open-ended user queries on our self-constructed dataset.
We evaluate the local Open-EA instantiation on four in-domain and three out-of-domain T2V generator families, analyze the dimensions on which its tier predictions remain difficult, and characterize its end-to-end operational cost separately from that of the API-based Evaluation Agent.

\subsection{Experiments on Existing Benchmarks}
\label{ssec:closed-domain}

We validate the effectiveness of our framework on both the Text-to-Video (T2V) and Text-to-Image (T2I) tasks. In this subsection, all closed-domain experiments use LLM agents with \texttt{gpt-4o-2024-08-06} as the backbone and a temperature of 0.7. The system prompts follow a CoT/ReAct-style protocol~\cite{wei2022chain, yao2022react}, requiring the agent to reason before selecting a tool, interpret intermediate observations, and justify its final response.

\subsubsection{Experimental Setup}

\noindent\textbf{Visual Generative Models.}
For the T2V task, we select four open-source models: VideoCrafter-0.9~\cite{he2022lvdm}, VideoCrafter-2~\cite{chen2024videocrafter2}, Latte-1~\cite{ma2024latte}, and ModelScope~\cite{wang2023modelscope}. Similarly, for the T2I task, we choose four well-known open-source models: SD(Stable Diffusion)1.4~\cite{rombach2022high}, SD2.1~\cite{rombach2022high}, SDXL~\cite{podell2023sdxl}, and SD3.0~\cite{esser2024scaling}. We validate the efficiency and accuracy of our method using these models within the evaluation frameworks for both T2I and T2V tasks.

\noindent\textbf{Visual Generation Benchmarks.}
For the T2V and T2I tasks, we adopt well-established and comprehensive evaluation frameworks tailored to their respective domains. Specifically, for the T2V task, we use VBench~\cite{huang2023vbench}, a comprehensive and fine-grained evaluation framework for video generation. For the T2I task, we select T2I-CompBench~\cite{huang2023t2icompbench}, which assesses the compositional generation capabilities of T2I models across multiple dimensions. We validate the effectiveness of our method using these evaluation frameworks.

\noindent\textbf{Comparison Setup.}
To ensure comparability and fairness in our experiments, we restrict the evaluation tools and prompts available to the Evaluation Agent. For T2V tasks, we used only the metrics and prompt lists from VBench, and for T2I tasks, only those from T2I-CompBench. For comparison, we categorize performance into five levels based on the performance density distribution. For VBench, we evaluate 15 dimensions and map raw metric outputs into five categorical tiers: \texttt{Very High}, \texttt{High}, \texttt{Moderate}, \texttt{Low}, and \texttt{Very Low}. For proportion-style dimensions, such as \texttt{Dynamic Degree}, the tiering thresholds are constructed from leaderboard statistics. For T2I-CompBench, we use four dimensions, \texttt{Color Binding}, \texttt{Shape Binding}, \texttt{Texture Binding}, and \texttt{Non-Spatial Relationships}; we exclude \texttt{Spatial Relationships} because its score distribution and statistical interpretation are not directly comparable under our small-sample setting.

\subsubsection{Results Analysis}
\label{sssec:t2v_results}

\noindent\textbf{API-Based Evaluation Cost.}
Table~\ref{tab:eval_cost_comparison} summarizes the evaluation-cost comparison between standard benchmarks and the API-based Evaluation Agent. Relative to the average cost of one standard benchmark dimension, the API-based Evaluation Agent reduces both wall-clock time and the number of generated samples across all evaluated T2V and T2I models.

\begin{table*}[t]
    \centering
    \small
    \setlength{\tabcolsep}{3pt}
    \caption{\textbf{Evaluation Cost Comparison with Standard Benchmarks.} We compare the original VBench and T2I-CompBench pipelines with the API-based Evaluation Agent. Each entry reports wall-clock evaluation time and the number of generated samples. Benchmark costs are shown both for the complete benchmark and as an average per evaluated dimension, while the Evaluation Agent cost corresponds to one user-requested dimension.}
    \resizebox{0.96\textwidth}{!}{%
    \begin{tabular}{c|c||c|c|c}
        \Xhline{1pt}
        \textbf{Task / Benchmark} & \textbf{Model} &
        \textbf{\Centerstack{Standard Benchmark\\Total Cost $\downarrow$}} &
        \textbf{\Centerstack{Standard Benchmark\\Avg. Cost per Dimension $\downarrow$}} &
        \textbf{\Centerstack{Evaluation Agent\\Cost per Query $\downarrow$}} \\
        \Xhline{1pt}
        \multirow{4}{*}{\Centerstack{T2V /\\VBench}}
        & Latte-1~\cite{ma2024latte} & 2557 min / 4355 samples & 170 min / 290 samples & 15 min / 25 samples \\
        & ModelScope~\cite{wang2023modelscope} & 1160 min / 4355 samples & 77 min / 290 samples & 6 min / 23 samples \\
        & VideoCrafter-0.9~\cite{he2022lvdm} & 1459 min / 4355 samples & 97 min / 290 samples & 9 min / 24 samples \\
        & VideoCrafter-2~\cite{chen2024videocrafter2} & 4261 min / 4355 samples & 284 min / 290 samples & 24 min / 23 samples \\
        \midrule
        \multirow{4}{*}{\Centerstack{T2I /\\T2I-CompBench}}
        & SD1.4~\cite{rombach2022high} & 563 min / 12000 samples & 141 min / 3000 samples & 5 min / 26 samples \\
        & SD2.1~\cite{rombach2022high} & 782 min / 12000 samples & 196 min / 3000 samples & 6 min / 26 samples \\
        & SDXL~\cite{podell2023sdxl} & 1543 min / 12000 samples & 386 min / 3000 samples & 8 min / 26 samples \\
        & SD3.0~\cite{esser2024scaling} & 1410 min / 12000 samples & 353 min / 3000 samples & 7 min / 25 samples \\
        \Xhline{1pt}
    \end{tabular}%
    }
    \label{tab:eval_cost_comparison}
    \vspace{-5pt}
\end{table*}

\noindent\textbf{Validation on VBench.} To highlight the efficiency of our proposed methods, we compare the time consumption and sample counts between VBench and our approach. As shown in Table~\ref{tab:eval_cost_comparison}, our method reduces evaluation time by more than $10\times$. Additionally, Table~\ref{tab:main_results_t2v} confirms the consistency of our evaluation results with VBench across various dimensions.
The quantitative results show that the Evaluation Agent achieves high prediction accuracy across most dimensions, demonstrating that our approach maintains accuracy while significantly reducing evaluation time.

\noindent\textbf{Validation on T2I-CompBench.}
\label{sssec:t2i_results}
We evaluate the Evaluation Agent's performance on T2I tasks, as shown in Table~\ref{tab:eval_cost_comparison}. Instead of requiring thousands of samples and several hours, the Evaluation Agent completes evaluations with just about 26 samples in five to eight minutes per dimension. Table~\ref{tab:main_results_t2i} compares the evaluation results, showcasing high accuracy within an error margin of one range.

\begin{figure}[t]
    \vspace{2pt}
    \centering
    \includegraphics[width=0.85\linewidth]{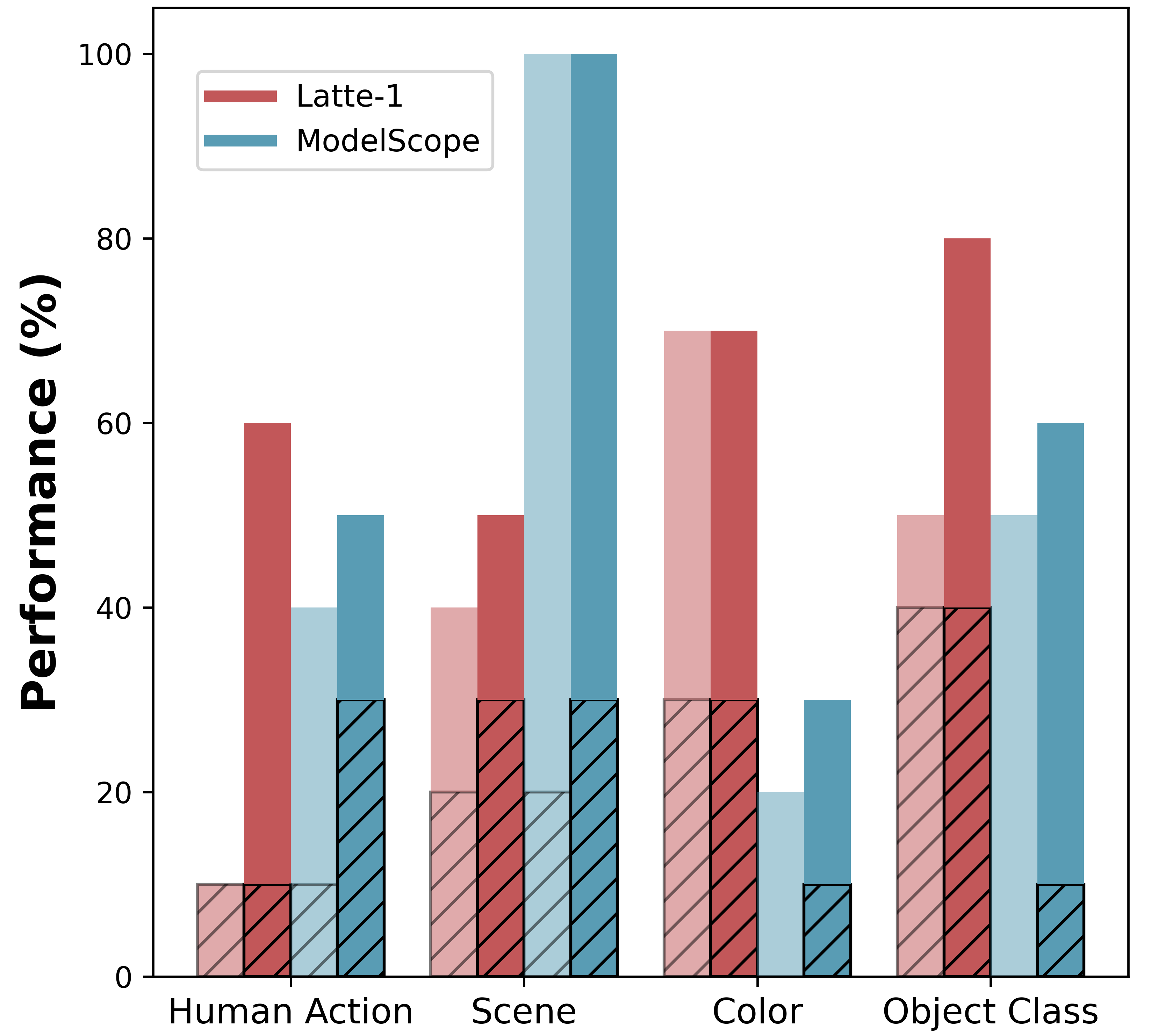}
    \caption{\textbf{Effect of Prompt Budget on Percentage-Based VBench Dimensions.} Lighter bars use the original prompt budget and darker bars use 30 prompts. Hatched portions denote exact-tier predictions and solid portions denote predictions within one neighboring tier.}
    \label{fig:t2v_percent_bar}
    \vspace{-10pt}
\end{figure}

\noindent\textbf{Effect of the Prompt Budget.}
\label{sssec:percentage_results}
The lower accuracy on Human Action, Scene, Color, and Object Class is consistent with their reliance on percentage estimates formed from binary sample-level outcomes. We therefore evaluate Latte-1 and ModelScope with 30 prompts instead of the default budget, which ranges from three to nine prompts per round. Figure~\ref{fig:t2v_percent_bar} shows that Latte-1 improves from 25.0\% to 27.5\% in Exact accuracy and from 42.5\% to 65.0\% in Within-1 accuracy. ModelScope improves from 7.5\% to 20.0\% in Exact accuracy and from 52.5\% to 60.0\% in Within-1 accuracy. When averaged across the two generators, Within-1 improves by 15.0 percentage points, which is twice the 7.5-point improvement in Exact accuracy. The larger gain under the one-tier tolerance suggests that additional prompts chiefly stabilize the estimated capability region, while the remaining Exact gap shows that a larger budget alone does not resolve errors near tier boundaries.

\noindent\textbf{Backbone Robustness of the API-Based Evaluation Agent.}
Keeping the evaluation protocol fixed, we replace \texttt{gpt-4o-2024-08-06} with \texttt{claude-3-5-sonnet-20241022}. Across all evaluated model--dimension pairs, GPT-4o and Claude obtain 44.8\% / 81.3\% versus 41.5\% / 77.2\% Exact / Within-1 accuracy on VBench, and 53.8\% / 96.3\% versus 54.4\% / 100.0\% on T2I-CompBench. Their close aggregate results indicate that the dynamic evaluation loop is not tied to GPT-4o, although individual dimensions differ. These experiments assess the API-based Evaluation Agent rather than EA-3B.

\subsection{Open-EA Quantitative Evaluation}
\label{ssec:open_ea_quantitative}

We evaluate EA-3B on all 15 VBench dimensions using four in-domain and three out-of-domain generators. The EA-CoT-10K training traces include Latte-1, ModelScope, VideoCrafter-0.9, and VideoCrafter-2, whereas Show-1 and the two CogVideoX variants do not appear in the training traces and therefore serve as out-of-domain generator families. Table~\ref{tab:open_ea_results} reports Exact and Within-1 tier accuracy over ten trials for every model--dimension pair.

\begin{table*}[t]
    \caption{\textbf{Open-EA Evaluation across VBench Dimensions.} Each entry reports Exact / Within-1 tier accuracy over ten trials. Four in-domain generators are evaluated alongside three out-of-domain families to measure cross-family transfer.}
    \vspace{-5pt}
    \centering
    \small
    \setlength{\tabcolsep}{2pt}
    \resizebox{\textwidth}{!}{%
        \begin{tabular}{>{\centering\arraybackslash}p{2.5cm}|
                            >{\centering\arraybackslash}p{2cm}|
                            >{\centering\arraybackslash}p{2cm}|
                            >{\centering\arraybackslash}p{2cm}|
                            >{\centering\arraybackslash}p{2cm}|
                            >{\centering\arraybackslash}p{2cm}|
                            >{\centering\arraybackslash}p{2cm}|
                            >{\centering\arraybackslash}p{2cm}}
        \Xhline{1pt}
        \textbf{Models} & \textbf{\Centerstack{Subject\\Consistency}} & \textbf{\Centerstack{Background\\Consistency}} &
        \textbf{\Centerstack{Motion\\Smoothness}} &
        \textbf{\Centerstack{Dynamic\\Degree}} &
        \textbf{\Centerstack{Aesthetic\\Quality}} &
        \textbf{\Centerstack{Imaging\\Quality}} &
        \textbf{\Centerstack{Object\\Class}} \\
        \Xhline{.8pt}
        \multicolumn{8}{l}{\textit{In-Domain Models}} \\
        \midrule
        Latte-1 & 60{\%} / 100{\%} & 40{\%} / 70{\%} & 90{\%} / 100{\%} & 80{\%} / 90{\%} & 30{\%} / 80{\%} & 40{\%} / 80{\%} & 0{\%} / 0{\%} \\
        ModelScope & 30{\%} / 50{\%} & 40{\%} / 60{\%} & 80{\%} / 90{\%} & 60{\%} / 80{\%} & 40{\%} / 90{\%} & 70{\%} / 100{\%} & 0{\%} / 30{\%} \\
        VideoCrafter-0.9 & 100{\%} / 100{\%} & 100{\%} / 100{\%} & 100{\%} / 100{\%} & 100{\%} / 100{\%} & 100{\%} / 100{\%} & 90{\%} / 100{\%} & 0{\%} / 10{\%} \\
        VideoCrafter-2 & 40{\%} / 100{\%} & 30{\%} / 100{\%} & 40{\%} / 80{\%} & 0{\%} / 100{\%} & 50{\%} / 100{\%} & 80{\%} / 100{\%} & 100{\%} / 100{\%} \\
        \midrule
        \multicolumn{8}{l}{\textit{Out-of-Domain Models}} \\
        \midrule
        Show-1 & 0{\%} / 40{\%} & 30{\%} / 80{\%} & 10{\%} / 60{\%} & 40{\%} / 80{\%} & 50{\%} / 100{\%} & 30{\%} / 100{\%} & 10{\%} / 30{\%} \\
        CogVideoX-2B & 70{\%} / 100{\%} & 50{\%} / 90{\%} & 80{\%} / 80{\%} & 0{\%} / 20{\%} & 50{\%} / 80{\%} & 20{\%} / 50{\%} & 60{\%} / 60{\%} \\
        CogVideoX-5B & 50{\%} / 80{\%} & 30{\%} / 80{\%} & 10{\%} / 90{\%} & 30{\%} / 80{\%} & 50{\%} / 70{\%} & 20{\%} / 70{\%} & 0{\%} / 20{\%} \\
        \Xhline{1pt}
        \end{tabular}
    }

    \vspace{5pt}

    \setlength{\tabcolsep}{2pt}
    \resizebox{\textwidth}{!}{%
    \begin{tabular}{>{\centering\arraybackslash}p{2.5cm}|
                            >{\centering\arraybackslash}p{1.8cm}|
                            >{\centering\arraybackslash}p{1.8cm}|
                            >{\centering\arraybackslash}p{1.8cm}|
                            >{\centering\arraybackslash}p{1.8cm}|
                            >{\centering\arraybackslash}p{1.8cm}|
                            >{\centering\arraybackslash}p{1.8cm}|
                            >{\centering\arraybackslash}p{1.8cm}|
                            >{\centering\arraybackslash}p{1.8cm}}
    \Xhline{1pt}
    \textbf{Models} &
    \textbf{\Centerstack{Multiple\\Objects}} &
    \textbf{\Centerstack{Human\\Action}} &
    \textbf{\Centerstack{Color}} &
    \textbf{\Centerstack{Spatial\\Relationship}} &
    \textbf{\Centerstack{Scene}} &
    \textbf{\Centerstack{Appearance\\Style}} &
    \textbf{\Centerstack{Temporal\\Style}} &
    \textbf{\Centerstack{Overall\\Consistency}} \\
    \Xhline{.8pt}
    \multicolumn{9}{l}{\textit{In-Domain Models}} \\
    \midrule
    Latte-1 & 20{\%} / 100{\%} & 0{\%} / 70{\%} & 0{\%} / 0{\%} & 0{\%} / 0{\%} & 80{\%} / 80{\%} & 40{\%} / 80{\%} & 60{\%} / 90{\%} & 20{\%} / 100{\%} \\
    ModelScope & 20{\%} / 100{\%} & 0{\%} / 0{\%} & 10{\%} / 30{\%} & 30{\%} / 30{\%} & 0{\%} / 70{\%} & 10{\%} / 100{\%} & 10{\%} / 60{\%} & 30{\%} / 100{\%} \\
    VideoCrafter-0.9 & 100{\%} / 100{\%} & 0{\%} / 0{\%} & 60{\%} / 80{\%} & 30{\%} / 100{\%} & 0{\%} / 100{\%} & 60{\%} / 100{\%} & 40{\%} / 100{\%} & 80{\%} / 100{\%} \\
    VideoCrafter-2 & 10{\%} / 80{\%} & 0{\%} / 60{\%} & 100{\%} / 100{\%} & 0{\%} / 60{\%} & 40{\%} / 40{\%} & 20{\%} / 100{\%} & 0{\%} / 100{\%} & 20{\%} / 100{\%} \\
    \midrule
    \multicolumn{9}{l}{\textit{Out-of-Domain Models}} \\
    \midrule
    Show-1 & 30{\%} / 80{\%} & 0{\%} / 60{\%} & 10{\%} / 40{\%} & 0{\%} / 40{\%} & 0{\%} / 0{\%} & 10{\%} / 60{\%} & 20{\%} / 90{\%} & 70{\%} / 100{\%} \\
    CogVideoX-2B & 40{\%} / 40{\%} & 60{\%} / 60{\%} & 20{\%} / 30{\%} & 0{\%} / 0{\%} & 60{\%} / 60{\%} & 30{\%} / 80{\%} & 20{\%} / 90{\%} & 10{\%} / 80{\%} \\
    CogVideoX-5B & 10{\%} / 30{\%} & 50{\%} / 50{\%} & 20{\%} / 100{\%} & 100{\%} / 100{\%} & 0{\%} / 0{\%} & 40{\%} / 70{\%} & 30{\%} / 100{\%} & 80{\%} / 100{\%} \\
    \Xhline{1pt}
    \end{tabular}
    }
    \label{tab:open_ea_results}
\end{table*}

\noindent\textbf{Generalization and Tier Calibration.}
Open-EA retains partial cross-family generalization: the micro-averaged Exact / Within-1 score decreases from 41.3\% / 77.3\% in domain to 31.1\% / 64.9\% out of domain. The persistent Exact--Within-1 gap suggests that coarse capability placement transfers more reliably than exact tier calibration; VideoCrafter-2 is the clearest example, with 35.3\% Exact but 88.0\% Within-1 accuracy.

\noindent\textbf{Difficulty across Evaluation Signals.}
Table~\ref{tab:open_ea_results} shows that performance varies sharply with the evaluation signal. Across all generators, Overall Consistency and Temporal Style achieve 97.1\% and 90.0\% Within-1 accuracy, whereas Object Class, Human Action, and Spatial Relationship reach only 35.7\%, 42.9\%, and 47.1\%, respectively. VideoCrafter-0.9 makes this contrast especially clear: it attains at least 80\% Within-1 accuracy on 13 of the 15 dimensions, yet scores 0\% / 0\% on Human Action and 0\% / 10\% on Object Class. A plausible explanation is that global quality and temporal properties provide repeated video-level evidence, while class-specific or relational dimensions infer a model-level tier from a small number of prompt-specific successes and are therefore more sensitive to concept coverage and tier boundaries. Together with the prompt-budget ablation in Figure~\ref{fig:t2v_percent_bar}, this pattern suggests that Open-EA transfers broad quality and consistency judgments more reliably than fine-grained semantic coverage, making prompt diversity and tier calibration the clearest opportunities for improvement.

\noindent\textbf{End-to-End Runtime.}
Table~\ref{tab:open_ea_speed} reports wall-clock time for the complete local Open-EA pipeline, rather than EA-3B latency alone.

\begin{table}[t]
    \centering
    \small
    \setlength{\tabcolsep}{3pt}
    \caption{\textbf{Open-EA End-to-End Inference Speed.} Wall-clock runtime over all successful saved traces, including planning, visual generation, tool execution, observation aggregation, and summarization. This trace pool is summarized independently of the fixed ten-trial accuracy evaluation.}
    \vspace{-5pt}
    \resizebox{0.85\linewidth}{!}{%
    \begin{tabular}{c|c|c|c}
        \Xhline{1pt}
        \textbf{Models} &
        \textbf{\Centerstack{Saved\\Runs}} &
        \textbf{\Centerstack{Median Time\\per Query $\downarrow$}} &
        \textbf{\Centerstack{Mean Time\\per Query $\downarrow$}} \\
        \Xhline{.8pt}
        \multicolumn{4}{l}{\textit{In-Domain Models with Saved Traces}} \\
        \midrule
        Latte-1 & 148 & 7.87 min & 9.23 min \\
        ModelScope & 147 & 1.86 min & 2.41 min \\
        \midrule
        \multicolumn{4}{l}{\textit{Out-of-Domain Models}} \\
        \midrule
        Show-1 & 117 & 144.35 min & 176.64 min \\
        CogVideoX-2B & 155 & 20.29 min & 23.89 min \\
        CogVideoX-5B & 149 & 52.42 min & 57.02 min \\
        \Xhline{1pt}
    \end{tabular}
    }
    \vspace{-5pt}
    \label{tab:open_ea_speed}
\end{table}

Median runtime spans 1.86--144.35 minutes, a 77.6$\times$ range despite the shared planning backbone, while consistently higher means indicate a right tail of slower runs. Because the saved traces differ in generator and tool mix, these measurements describe full-pipeline deployment cost; they cannot rank EA-3B planning speed or be compared directly with the API-based Evaluation Agent costs in Table~\ref{tab:eval_cost_comparison}.

\subsection{Experiments on Open-Ended User Query}

We demonstrate the flexibility of our framework and the benefits of its dynamic evaluation through experiments on an open-ended user query dataset that we collected and curated.
Unless otherwise stated, the cases in this subsection use the API-based Evaluation Agent rather than EA-3B.

\subsubsection{Open-Ended User Query Dataset}

We create an open-ended user query dataset comprising 100 user queries focused on evaluating generative model capabilities. Each query is manually labeled with \texttt{Ability}, \texttt{General/Specific}, and \texttt{Specific Domain} tags.
The dataset is collected through a user study and then cleaned, filtered, and expanded into 100 final queries. The ability labels cover \texttt{Prompt Following}, \texttt{Visual Quality}, \texttt{Creativity}, \texttt{Knowledge}, and \texttt{Others}; the scope labels distinguish general questions from application-specific questions; and the domain labels cover areas such as law, film and entertainment, fashion, game design, architecture and interior design, medical, science and education, and history and culture. In the current dataset, 60 queries are general and 40 are specific, and the ability categories are distributed across knowledge, visual quality, creativity, prompt following, and other model capabilities.

\begin{figure*}[htbp!]
    \centering
    \includegraphics[width=0.9\textwidth]{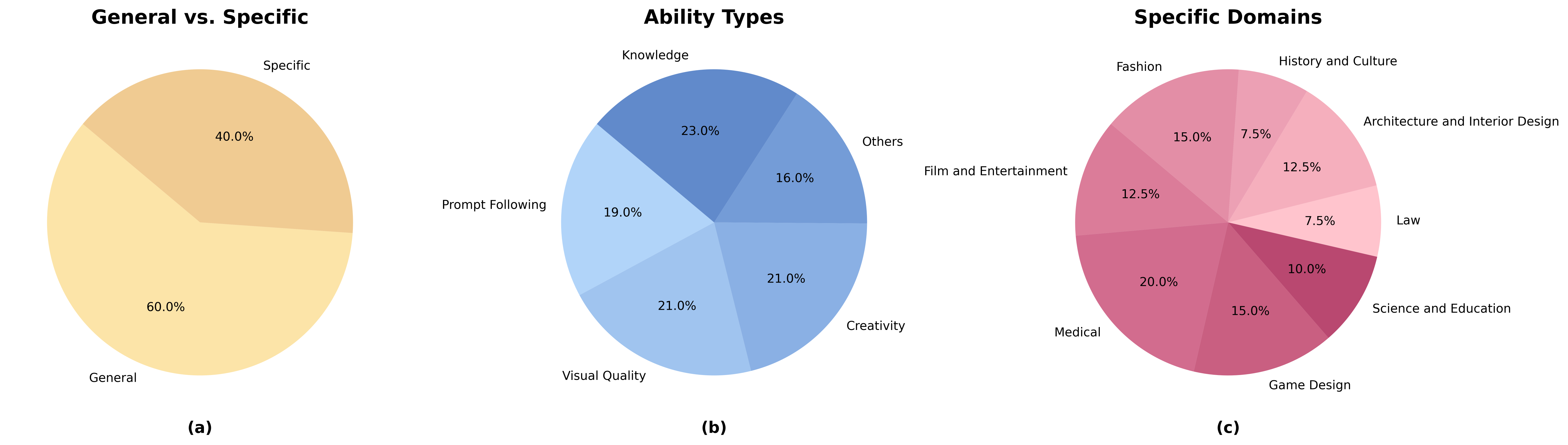}
    \caption{\textbf{Data Distribution of Open-Ended User Query Dataset.} We analyze the constructed open-ended user query dataset from three aspects: General/Specific, Ability, and Specific Domain. The distribution shows that the collected queries cover both broad capability questions and application-oriented use cases.}
    \label{fig:open_domain_stat}
    \vspace{-5pt}
\end{figure*}

\subsubsection{Experimental Setup}
The Evaluation Agent demonstrates strong planning capabilities and accepts any input, but its effectiveness is limited by restrictive evaluation tools, which hinder its ability to handle open-ended queries. To overcome this limitation, we propose a simple yet intuitive solution: leveraging a VLM as an evaluation tool in the form of VQA. During each evaluation round, the PromptGen Agent not only designs prompts for specific sub-aspects but also generates corresponding questions based on the content of each prompt and the aspects to be evaluated. The generated sample and questions are input into the VLM, which provides answers that are then fed back to the Plan Agent for further analysis and planning.
This VQA formulation converts open-ended visual inspection into a structured textual observation. As a result, the Plan Agent can compare evidence across rounds, decide whether to broaden the tested scenarios or increase prompt complexity, and produce a final answer grounded in the generated samples rather than in a single scalar metric.

\subsubsection{Open-Ended User Query Evaluation}

Most visual generation benchmarks use predefined dimensions and prompts to evaluate models, but this fixed approach often overlooks users' specific needs, such as handling unique scenarios or objects. A user study found that 67.44\% (29 of 43) participants prioritized models meeting their specific needs over general performance. Additionally, fixed prompts can lead to targeted optimization, resulting in misleading evaluations.

\begin{figure}[t]
    \centering
    \includegraphics[width=0.9\linewidth]{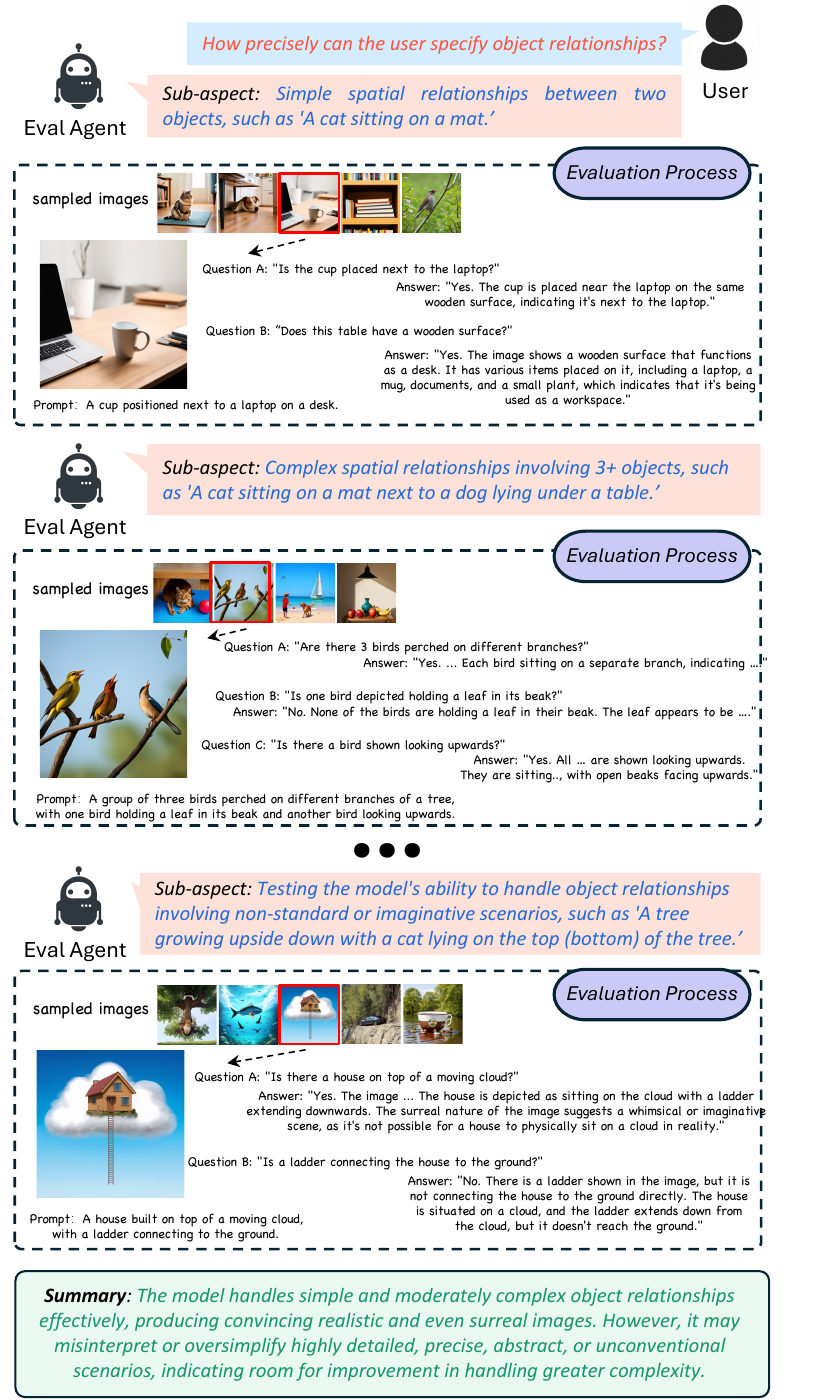}
    \caption{\textbf{A Complete Case of Open-Ended User Query Evaluation.} Given the query ``How precisely can the user specify object relationships?'', the Evaluation Agent progressively probes simple two-object relations, complex relations among three or more objects, and non-standard imaginative configurations. Round-wise VQA observations guide subsequent probes and support the final conclusion: the evaluated model handles simple and moderately complex relations but remains less reliable under highly detailed or unconventional constraints.}
    \label{fig:open_case}
    \vspace{-5pt}
\end{figure}

We address these issues using the Evaluation Agent, which conducts dynamic, multi-round assessments of a model's capabilities for open-ended queries, with flexible prompt design at each stage. Figure~\ref{fig:open_case} presents a complete evaluation trajectory for the query: ``How precisely can the user specify object relationships?'' The agent first evaluates simple spatial relationships between two objects, then increases the difficulty to relationships involving three or more objects, and finally probes non-standard and imaginative configurations. The intermediate VQA observations show that the evaluated model captures common spatial relationships but may omit fine-grained constraints, such as a leaf held in a bird's beak or a ladder reaching the ground. By adapting its focus across rounds and synthesizing the accumulated evidence, the agent distinguishes reliable performance on simple and moderately complex relationships from persistent limitations on highly detailed or unconventional arrangements. Additional evaluated queries include artistic-style preservation and specific object-count generation, where the agent records the sub-aspects, thoughts, observations, and final conclusions in the same trajectory format. As shown, our Evaluation Agent framework provides precise, detailed, and user-focused evaluations for open-ended queries.

\section{Further Discussions}

In this section, we discuss the unique aspects of the Evaluation Agent compared to traditional benchmarks, as well as its potential broader applications.

\noindent\textbf{Dynamic and Multi-Step Evaluation.} One of the core features of the Evaluation Agent is its dynamic, multi-step evaluation process. This structured evaluation paradigm enables the discovery of nuanced differences in model capabilities and provides a more detailed analysis of a model's strengths and weaknesses. Specifically, it allows for a hierarchical, step-by-step evaluation, progressing from simple to complex tasks, as well as category-based assessments that measure performance across different content types within the same dimension. In contrast, traditional visual generation evaluation benchmarks, while incorporating diverse prompts carefully designed for each dimension, suffer from limitations such as fixed prompts and a lack of fine-grained prompt categorization. These constraints reduce flexibility, make it harder to draw valuable insights, and increase the risk of models being over-optimized for specific prompts. Furthermore, dynamic evaluations help avoid redundant testing, significantly enhancing efficiency.

\noindent\textbf{Open-Ended Evaluation Toolkit.} Our framework for evaluating open-ended queries relies on two key aspects: the planning and reasoning capabilities of LLM Agents and the evaluation toolkit's ability to assess diverse dimensions. One approach to building the evaluation toolkit is integrating various tools, allowing the agent to select the most suitable one for each task. However, current evaluation tools are limited, often focusing only on general evaluations and lacking sensitivity to fine-grained details. For instance, CLIPScore~\cite{hessel2021clipscore} captures the general similarity between an image and a caption but fails to detect subtle changes, such as variations in object counts, limiting its effectiveness in specific evaluations. An alternative approach is to design a versatile evaluation tool capable of assessing multiple aspects. A VQA-based format using VLMs is particularly promising, enabling fine-grained evaluation through targeted questions and providing detailed textual outputs that integrate well with LLM Agents. While this method's effectiveness depends on the VLM's capabilities, current VLMs already demonstrate impressive results, with future advancements poised to further enhance performance.

\noindent\textbf{Agent Reliability and Error Handling.} Although the Plan Agent follows a structured multi-step protocol, agentic evaluation can still fail through unreasonable tool choices, repetitive proposals, or premature/non-terminating decisions. We therefore combine prompt-level safeguards, such as requiring an explicit thought before each action and before termination, with system-level constraints such as a maximum number of rounds. These mechanisms do not remove the need for stronger reliability design, but they make the evaluation trace auditable and prevent unbounded evaluation cost.

\noindent\textbf{Broader Applications.} The Evaluation Agent not only evaluates the performance of a single model but also facilitates the direct comparison of two models' strengths and weaknesses in specific dimensions by assessing them simultaneously during execution. This feature enables users to determine which model excels in particular areas. Moreover, by accumulating evaluation results across various capabilities, a database can be constructed. Once sufficient information about multiple models is collected, this database can serve as the foundation for building a recommendation system, capable of suggesting the most suitable model based on the user's specific needs.

\section{Conclusion}
\label{conclusion}

We present Evaluation Agent, a dynamic and promptable framework for assessing visual generative models beyond rigid evaluation pipelines. Unlike traditional methods that rely on fixed benchmarks and time-consuming sampling processes, the Evaluation Agent adapts its evaluation trajectory to intermediate evidence and user-specified criteria. This design reduces the number of required samples while supporting flexible integration of evaluation tools and visual generative models.
By open-sourcing this framework, we aim to inspire further research in the development of more flexible and efficient evaluation methods for visual generative models.
The Open-EA results show that this protocol can be distilled into a compact local agent trained from evaluation trajectories, reducing dependence on proprietary planning backbones while preserving promptable, tool-grounded evaluation.

\section{Acknowledgment}
This study is supported by the Ministry of Education, Singapore, under its MOE AcRF Tier 2 (MOET2EP20221-0012, MOE-T2EP20223-0002), and under the RIE2020 Industry Alignment Fund -- Industry Collaboration Projects (IAF-ICP) Funding Initiative, as well as cash and in-kind contribution from the industry partner(s).

\bibliographystyle{IEEEtran}
\bibliography{ref}

@inproceedings{Jain_2023,
   title={Multi-Dimensional Evaluation of Text Summarization with In-Context Learning},
   url={http://dx.doi.org/10.18653/v1/2023.findings-acl.537},
   DOI={10.18653/v1/2023.findings-acl.537},
   booktitle={Findings of the Association for Computational Linguistics: ACL 2023},
   publisher={Association for Computational Linguistics},
   author={Jain, Sameer and Keshava, Vaishakh and Mysore Sathyendra, Swarnashree and Fernandes, Patrick and Liu, Pengfei and Neubig, Graham and Zhou, Chunting},
   year={2023},
   pages={8487--8495} }

@inproceedings{chiang2023largelanguagemodelsalternative,
      title={Can Large Language Models Be an Alternative to Human Evaluations?},
      author={Cheng-Han Chiang and Hung-yi Lee},
      booktitle={Proceedings of the 61st Annual Meeting of the Association for Computational Linguistics},
      year={2023},
      url={https://arxiv.org/abs/2305.01937},
}

@inproceedings{fu2023gptscoreevaluatedesire,
      title={GPTScore: Evaluate as You Desire},
      author={Fu, Jinlan and Ng, See-Kiong and Jiang, Zhengbao and Liu, Pengfei},
      booktitle={Proceedings of the 2024 Conference of the North American Chapter of the Association for Computational Linguistics: Human Language Technologies},
      pages={6556--6576},
      year={2024},
      doi={10.18653/v1/2024.naacl-long.365},
      url={https://aclanthology.org/2024.naacl-long.365/},
}

@inproceedings{li2023collaborativeevaluationexploringsynergy,
      title={Exploring the Reliability of Large Language Models as Customized Evaluators for Diverse NLP Tasks},
      author={Li, Qintong and Cui, Leyang and Kong, Lingpeng and Bi, Wei},
      booktitle={Proceedings of the 31st International Conference on Computational Linguistics},
      pages={10325--10344},
      year={2025},
      url={https://aclanthology.org/2025.coling-main.688/},
}

@inproceedings{pan2024autonomous,
  title={Autonomous evaluation and refinement of digital agents},
  author={Pan, Jiayi and Zhang, Yichi and Tomlin, Nicholas and Zhou, Yifei and Levine, Sergey and Suhr, Alane},
  booktitle={Proceedings of the First Conference on Language Modeling},
  year={2024}
}

@inproceedings{zhang2025evaluationagent,
  title={{Evaluation Agent}: Efficient and Promptable Evaluation Framework for Visual Generative Models},
  author={Zhang, Fan and Tian, Shulin and Huang, Ziqi and Qiao, Yu and Liu, Ziwei},
  booktitle={Proceedings of the 63rd Annual Meeting of the Association for Computational Linguistics (Volume 1: Long Papers)},
  pages={7561--7582},
  year={2025},
  publisher={Association for Computational Linguistics},
  doi={10.18653/v1/2025.acl-long.374}
}

@article{wang2024survey,
  title={A survey on large language model based autonomous agents},
  author={Wang, Lei and Ma, Chen and Feng, Xueyang and Zhang, Zeyu and Yang, Hao and Zhang, Jingsen and Chen, Zhiyuan and Tang, Jiakai and Chen, Xu and Lin, Yankai and others},
  journal={Frontiers of Computer Science},
  volume={18},
  number={6},
  pages={186345},
  year={2024},
  publisher={Springer}
}

@inproceedings{yao2022react,
  title={React: Synergizing reasoning and acting in language models},
  author={Yao, Shunyu and Zhao, Jeffrey and Yu, Dian and Du, Nan and Shafran, Izhak and Narasimhan, Karthik and Cao, Yuan},
  booktitle={International Conference on Learning Representations},
  year={2023}
}

@article{shinn2024reflexion,
  title={Reflexion: Language agents with verbal reinforcement learning},
  author={Shinn, Noah and Cassano, Federico and Gopinath, Ashwin and Narasimhan, Karthik and Yao, Shunyu},
  journal={Advances in Neural Information Processing Systems},
  volume={36},
  year={2024}
}

@article{wei2022chain,
  title={Chain-of-thought prompting elicits reasoning in large language models},
  author={Wei, Jason and Wang, Xuezhi and Schuurmans, Dale and Bosma, Maarten and Xia, Fei and Chi, Ed and Le, Quoc V and Zhou, Denny and others},
  journal={Advances in neural information processing systems},
  volume={35},
  pages={24824--24837},
  year={2022}
}

@article{kojima2022large,
  title={Large language models are zero-shot reasoners},
  author={Kojima, Takeshi and Gu, Shixiang Shane and Reid, Machel and Matsuo, Yutaka and Iwasawa, Yusuke},
  journal={Advances in neural information processing systems},
  volume={35},
  pages={22199--22213},
  year={2022}
}

@article{yao2024tree,
  title={Tree of thoughts: Deliberate problem solving with large language models},
  author={Yao, Shunyu and Yu, Dian and Zhao, Jeffrey and Shafran, Izhak and Griffiths, Tom and Cao, Yuan and Narasimhan, Karthik},
  journal={Advances in Neural Information Processing Systems},
  volume={36},
  year={2024}
}

@inproceedings{sel2023algorithm,
  title={Algorithm of thoughts: Enhancing exploration of ideas in large language models},
  author={Sel, Bilgehan and Al-Tawaha, Ahmad and Khattar, Vanshaj and Jia, Ruoxi and Jin, Ming},
  booktitle={International Conference on Machine Learning},
  year={2024}
}

@inproceedings{besta2024graph,
  title={Graph of thoughts: Solving elaborate problems with large language models},
  author={Besta, Maciej and Blach, Nils and Kubicek, Ales and Gerstenberger, Robert and Podstawski, Michal and Gianinazzi, Lukas and Gajda, Joanna and Lehmann, Tomasz and Niewiadomski, Hubert and Nyczyk, Piotr and others},
  booktitle={Proceedings of the AAAI Conference on Artificial Intelligence},
  volume={38},
  number={16},
  pages={17682--17690},
  year={2024}
}

@inproceedings{wang2022self,
  title={Self-consistency improves chain of thought reasoning in language models},
  author={Wang, Xuezhi and Wei, Jason and Schuurmans, Dale and Le, Quoc and Chi, Ed and Narang, Sharan and Chowdhery, Aakanksha and Zhou, Denny},
  booktitle={International Conference on Learning Representations},
  year={2023}
}

@inproceedings{li2023api,
  title={Api-bank: A comprehensive benchmark for tool-augmented llms},
  author={Li, Minghao and Zhao, Yingxiu and Yu, Bowen and Song, Feifan and Li, Hangyu and Yu, Haiyang and Li, Zhoujun and Huang, Fei and Li, Yongbin},
  booktitle={Proceedings of the 2023 Conference on Empirical Methods in Natural Language Processing},
  pages={3102--3116},
  year={2023},
  doi={10.18653/v1/2023.emnlp-main.187}
}

@inproceedings{qin2023toolllm,
  title={Toolllm: Facilitating large language models to master 16000+ real-world apis},
  author={Qin, Yujia and Liang, Shihao and Ye, Yining and Zhu, Kunlun and Yan, Lan and Lu, Yaxi and Lin, Yankai and Cong, Xin and Tang, Xiangru and Qian, Bill and others},
  booktitle={International Conference on Learning Representations},
  year={2024}
}

@article{schick2024toolformer,
  title={Toolformer: Language models can teach themselves to use tools},
  author={Schick, Timo and Dwivedi-Yu, Jane and Dess{\`\i}, Roberto and Raileanu, Roberta and Lomeli, Maria and Hambro, Eric and Zettlemoyer, Luke and Cancedda, Nicola and Scialom, Thomas},
  journal={Advances in Neural Information Processing Systems},
  volume={36},
  year={2024}
}

@inproceedings{patil2023gorillalargelanguagemodel,
      title={Gorilla: Large Language Model Connected with Massive APIs},
      author={Patil, Shishir G. and Zhang, Tianjun and Wang, Xin and Gonzalez, Joseph E.},
      booktitle={Advances in Neural Information Processing Systems},
      volume={37},
      year={2024},
      doi={10.52202/079017-4020},
}

@inproceedings{ho2020denoising,
  title={Denoising Diffusion Probabilistic Models},
  author={Jonathan Ho and Ajay Jain and Pieter Abbeel},
  year={2020},
  booktitle={Advances in Neural Information Processing Systems},
  volume={33}
}

@InProceedings{Bain21,
  author       = "Max Bain and Arsha Nagrani and G{\"u}l Varol and Andrew Zisserman",
  title        = "Frozen in Time: A Joint Video and Image Encoder for End-to-End Retrieval",
  booktitle    = "IEEE International Conference on Computer Vision",
  year         = "2021",
}

@inproceedings{chen2024panda70m,
    title   = {Panda-70M: Captioning 70M Videos with Multiple Cross-Modality Teachers},
    author  = {Chen, Tsai-Shien and Siarohin, Aliaksandr and Menapace, Willi and Deyneka, Ekaterina and Chao, Hsiang-wei and Jeon, Byung Eun and Fang, Yuwei and Lee, Hsin-Ying and Ren, Jian and Yang, Ming-Hsuan and Tulyakov, Sergey},
    booktitle = {Proceedings of the IEEE/CVF Conference on Computer Vision and Pattern Recognition},
    year    = {2024}
}

@article{xue2019video,
  title={Video Enhancement with Task-Oriented Flow},
  author={Xue, Tianfan and Chen, Baian and Wu, Jiajun and Wei, Donglai and Freeman, William T},
  journal={International Journal of Computer Vision (IJCV)},
  volume={127},
  number={8},
  pages={1106--1125},
  year={2019},
  publisher={Springer}
}

@InProceedings{huang2023vbench,
     title={{VBench}: Comprehensive Benchmark Suite for Video Generative Models},
     author={Huang, Ziqi and He, Yinan and Yu, Jiashuo and Zhang, Fan and Si, Chenyang and Jiang, Yuming and Zhang, Yuanhan and Wu, Tianxing and Jin, Qingyang and Chanpaisit, Nattapol and Wang, Yaohui and Chen, Xinyuan and Wang, Limin and Lin, Dahua and Qiao, Yu and Liu, Ziwei},
     booktitle={Proceedings of the IEEE/CVF Conference on Computer Vision and Pattern Recognition},
     year={2024}
 }

@inproceedings{liu2023evalcrafter,
title={Evalcrafter: Benchmarking and evaluating large video generation models},
author={Liu, Yaofang and Cun, Xiaodong and Liu, Xuebo and Wang, Xintao and Zhang, Yong and Chen, Haoxin and Liu, Yang and Zeng, Tieyong and Chan, Raymond and Shan, Ying},
booktitle={Proceedings of the IEEE/CVF Conference on Computer Vision and Pattern Recognition},
pages={22139--22149},
year={2024},
doi={10.1109/CVPR52733.2024.02090}
}

@inproceedings{huang2023t2icompbench,
      title={T2I-CompBench: A Comprehensive Benchmark for Open-world Compositional Text-to-image Generation},
      author={Kaiyi Huang and Kaiyue Sun and Enze Xie and Zhenguo Li and Xihui Liu},
      booktitle={Advances in Neural Information Processing Systems},
      volume={36},
      year={2023},
}

@article{unterthiner2018towards,
  title={Towards accurate generative models of video: A new metric \& challenges},
  author={Unterthiner, Thomas and Van Steenkiste, Sjoerd and Kurach, Karol and Marinier, Raphael and Michalski, Marcin and Gelly, Sylvain},
  journal={arXiv preprint arXiv:1812.01717},
  year={2018}
}

@article{heusel2017gans,
  title={Gans trained by a two time-scale update rule converge to a local nash equilibrium},
  author={Heusel, Martin and Ramsauer, Hubert and Unterthiner, Thomas and Nessler, Bernhard and Hochreiter, Sepp},
  journal={Advances in neural information processing systems},
  volume={30},
  year={2017}
}

@article{ma2024latte,
  title={Latte: Latent diffusion transformer for video generation},
  author={Ma, Xin and Wang, Yaohui and Jia, Gengyun and Chen, Xinyuan and Liu, Ziwei and Li, Yuan-Fang and Chen, Cunjian and Qiao, Yu},
  journal={Transactions on Machine Learning Research},
  year={2025}
}

@article{wang2025lavie,
  title={Lavie: High-quality video generation with cascaded latent diffusion models},
  author={Wang, Yaohui and Chen, Xinyuan and Ma, Xin and Zhou, Shangchen and Huang, Ziqi and Wang, Yi and Yang, Ceyuan and He, Yinan and Yu, Jiashuo and Yang, Peiqing and others},
  journal={International Journal of Computer Vision},
  volume={133},
  number={5},
  pages={3059--3078},
  year={2025},
  publisher={Springer}
}

@inproceedings{chen2024videocrafter2,
  title={Videocrafter2: Overcoming data limitations for high-quality video diffusion models},
  author={Chen, Haoxin and Zhang, Yong and Cun, Xiaodong and Xia, Menghan and Wang, Xintao and Weng, Chao and Shan, Ying},
  booktitle={Proceedings of the IEEE/CVF Conference on Computer Vision and Pattern Recognition},
  pages={7310--7320},
  year={2024},
  doi={10.1109/CVPR52733.2024.00698}
}

@article{wang2023modelscope,
  title={Modelscope text-to-video technical report},
  author={Wang, Jiuniu and Yuan, Hangjie and Chen, Dayou and Zhang, Yingya and Wang, Xiang and Zhang, Shiwei},
  journal={arXiv preprint arXiv:2308.06571},
  year={2023}
}

@article{he2022lvdm,
  author={He, Yingqing and Yang, Tianyu and Zhang, Yong and Shan, Ying and Chen, Qifeng},
  journal={arXiv preprint arXiv:2211.13221},
  title={Latent video diffusion models for high-fidelity video generation with arbitrary lengths},
  year={2022}
}

@inproceedings{rombach2022high,
  title={High-resolution image synthesis with latent diffusion models},
  author={Rombach, Robin and Blattmann, Andreas and Lorenz, Dominik and Esser, Patrick and Ommer, Bj{\"o}rn},
  booktitle={Proceedings of the IEEE/CVF conference on computer vision and pattern recognition},
  pages={10684--10695},
  year={2022}
}

@inproceedings{podell2023sdxl,
  title={Sdxl: Improving latent diffusion models for high-resolution image synthesis},
  author={Podell, Dustin and English, Zion and Lacey, Kyle and Blattmann, Andreas and Dockhorn, Tim and M{\"u}ller, Jonas and Penna, Joe and Rombach, Robin},
  booktitle={International Conference on Learning Representations},
  year={2024}
}

@inproceedings{esser2024scaling,
  title={Scaling rectified flow transformers for high-resolution image synthesis},
  author={Esser, Patrick and Kulal, Sumith and Blattmann, Andreas and Entezari, Rahim and M{\"u}ller, Jonas and Saini, Harry and Levi, Yam and Lorenz, Dominik and Sauer, Axel and Boesel, Frederic and others},
  booktitle={Forty-first International Conference on Machine Learning},
  year={2024}
}

@article{jiang2026t2i,
  title={T2i-r1: Reinforcing image generation with collaborative semantic-level and token-level cot},
  author={Jiang, Dongzhi and Guo, Ziyu and Zhang, Renrui and Zong, Zhuofan and Li, Hao and Zhuo, Le and Yan, Shilin and Heng, Pheng-Ann and Li, Hongsheng},
  journal={Advances in Neural Information Processing Systems},
  volume={38},
  pages={39856--39890},
  year={2026}
}

@InProceedings{huang2026vchain,
        title={{VChain}: Chain-of-Visual-Thought for Reasoning in Video Generation},
        author = {Huang, Ziqi and Yu, Ning and Chen, Gordon and Qiu, Haonan and Debevec, Paul and Liu, Ziwei},
        booktitle={Annual Meeting of the Association for Computational Linguistics (ACL Findings), 2026},
        year={2026}
      }

@article{zheng2023judging,
  title={Judging llm-as-a-judge with mt-bench and chatbot arena},
  author={Zheng, Lianmin and Chiang, Wei-Lin and Sheng, Ying and Zhuang, Siyuan and Wu, Zhanghao and Zhuang, Yonghao and Lin, Zi and Li, Zhuohan and Li, Dacheng and Xing, Eric and others},
  journal={Advances in Neural Information Processing Systems},
  volume={36},
  pages={46595--46623},
  year={2023}
}

@inproceedings{xie2024osworld,
  title={Osworld: Benchmarking multimodal agents for open-ended tasks in real computer environments},
  author={Xie, Tianbao and Zhang, Danyang and Chen, Jixuan and Li, Xiaochuan and Zhao, Siheng and Cao, Ruisheng and Hua, Toh Jing and Cheng, Zhoujun and Shin, Dongchan and Lei, Fangyu and others},
  booktitle={Advances in Neural Information Processing Systems},
  volume={37},
  year={2024}
}

@inproceedings{zhou2023webarena,
  title={Webarena: A realistic web environment for building autonomous agents},
  author={Zhou, Shuyan and Xu, Frank F and Zhu, Hao and Zhou, Xuhui and Lo, Robert and Sridhar, Abishek and Cheng, Xianyi and Ou, Tianyue and Bisk, Yonatan and Fried, Daniel and others},
  booktitle={Advances in Neural Information Processing Systems},
  year={2024}
}

@inproceedings{wang2024mobile,
  title={Mobile-agent: Autonomous multi-modal mobile device agent with visual perception},
  author={Wang, Junyang and Xu, Haiyang and Ye, Jiabo and Yan, Ming and Shen, Weizhou and Zhang, Ji and Huang, Fei and Sang, Jitao},
  booktitle={ICLR 2024 Workshop on Large Language Model Agents},
  year={2024}
}

@inproceedings{kapoor2024omniact,
  title={OmniACT: A Dataset and Benchmark for Enabling Multimodal Generalist Autonomous Agents for Desktop and Web},
  author={Kapoor, Raghav and Butala, Yash Parag and Russak, Melisa and Koh, Jing Yu and Kamble, Kiran and Alshikh, Waseem and Salakhutdinov, Ruslan},
  booktitle={Computer Vision -- ECCV 2024},
  pages={161--178},
  year={2024},
  doi={10.1007/978-3-031-73113-6_10}
}

@inproceedings{tian2025mmina,
  title={{MMInA}: Benchmarking Multihop Multimodal Internet Agents},
  author={Tian, Shulin and Zhang, Ziniu and Chen, Liangyu and Liu, Ziwei},
  booktitle={Findings of the Association for Computational Linguistics: ACL 2025},
  pages={13682--13697},
  year={2025},
  doi={10.18653/v1/2025.findings-acl.703}
}

@article{huang2024vbench++,
  title={VBench++: Comprehensive and Versatile Benchmark Suite for Video Generative Models},
  author={Huang, Ziqi and Zhang, Fan and Xu, Xiaojie and He, Yinan and Yu, Jiashuo and Dong, Ziyue and Ma, Qianli and Chanpaisit, Nattapol and Si, Chenyang and Jiang, Yuming and Wang, Yaohui and Chen, Xinyuan and Chen, Ying-Cong and Wang, Limin and Lin, Dahua and Qiao, Yu and Liu, Ziwei},
  journal={IEEE Transactions on Pattern Analysis and Machine Intelligence},
  volume={48},
  number={3},
  pages={3268--3285},
  year={2026},
  doi={10.1109/TPAMI.2025.3633890}
}

@inproceedings{hessel2021clipscore,
  title={Clipscore: A reference-free evaluation metric for image captioning},
  author={Hessel, Jack and Holtzman, Ari and Forbes, Maxwell and Bras, Ronan Le and Choi, Yejin},
  booktitle={Proceedings of the 2021 Conference on Empirical Methods in Natural Language Processing},
  year={2021}
}

@article{qin2024evaluating,
  title={Evaluating Hallucination in Text-to-Image Diffusion Models with Scene-Graph based Question-Answering Agent},
  author={Qin, Ziyuan and Cheng, Dongjie and Wang, Haoyu and Yi, Huahui and Shao, Yuting and Fan, Zhiyuan and Li, Kang and Lao, Qicheng},
  journal={arXiv preprint arXiv:2412.05722},
  year={2024}
}

@article{zheng2025vbench2,
     title={{VBench-2.0}: Advancing Video Generation Benchmark Suite for Intrinsic Faithfulness},
     author={Zheng, Dian and Huang, Ziqi and Liu, Hongbo and Zou, Kai and He, Yinan and Zhang, Fan and Zhang, Yuanhan and He, Jingwen and Zheng, Wei-Shi and Qiao, Yu and Liu, Ziwei},
     journal={arXiv preprint arXiv:2503.21755},
     year={2025}
 }

@article{yang2025videogen,
  title={{VideoGen-Eval}: Agent-based System for Video Generation Evaluation},
  author={Yang, Yuhang and Fan, Ke and Sun, Shangkun and Li, Hongxiang and Zeng, Ailing and Han, FeiLin and Zhai, Wei and Liu, Wei and Cao, Yang and Zha, Zheng-Jun},
  journal={arXiv preprint arXiv:2503.23452},
  year={2025}
}

@article{lee2023holistic,
  title={Holistic evaluation of text-to-image models},
  author={Lee, Tony and Yasunaga, Michihiro and Meng, Chenlin and Mai, Yifan and Park, Joon Sung and Gupta, Agrim and Zhang, Yunzhi and Narayanan, Deepak and Teufel, Hannah and Bellagente, Marco and others},
  journal={Advances in Neural Information Processing Systems},
  volume={36},
  pages={69981--70011},
  year={2023}
}

@inproceedings{sun2024t2v,
  title={T2v-compbench: A comprehensive benchmark for compositional text-to-video generation},
  author={Sun, Kaiyue and Huang, Kaiyi and Liu, Xian and Wu, Yue and Xu, Zihan and Li, Zhenguo and Liu, Xihui},
  booktitle={Proceedings of the IEEE/CVF Conference on Computer Vision and Pattern Recognition},
  pages={8406--8416},
  year={2025},
  doi={10.1109/CVPR52734.2025.00787}
}

@article{yang2024qwen2technicalreport,
  title={Qwen2.5 Technical Report},
  author={{Qwen Team}},
  journal={arXiv preprint arXiv:2412.15115},
  year={2024}
}

@inproceedings{zheng2024llamafactory,
  title={LlamaFactory: Unified Efficient Fine-Tuning of 100+ Language Models},
  author={Zheng, Yaowei and Zhang, Richong and Zhang, Junhao and Ye, Yanhan and Luo, Zheyan},
  booktitle={Proceedings of the 62nd Annual Meeting of the Association for Computational Linguistics (Volume 3: System Demonstrations)},
  pages={400--410},
  year={2024},
  doi={10.18653/v1/2024.acl-demos.38}
}

@inproceedings{zhuge2025agentasjudge,
  title={{Agent-as-a-Judge}: Evaluate Agents with Agents},
  author={Zhuge, Mingchen and Zhao, Changsheng and Ashley, Dylan R. and Wang, Wenyi and Khizbullin, Dmitrii and Xiong, Yunyang and Liu, Zechun and Chang, Ernie and Krishnamoorthi, Raghuraman and Tian, Yuandong and Shi, Yangyang and Chandra, Vikas and Schmidhuber, J{\"u}rgen},
  booktitle={Proceedings of the 42nd International Conference on Machine Learning},
  series={Proceedings of Machine Learning Research},
  volume={267},
  pages={80569--80611},
  year={2025}
}

@inproceedings{brown2025taskelicitation,
  title={Adaptively Profiling Models with Task Elicitation},
  author={Brown, Davis and Balehannina, Prithvi and Jin, Helen and Havaldar, Shreya and Hassani, Hamed and Wong, Eric},
  booktitle={Proceedings of the 2025 Conference on Empirical Methods in Natural Language Processing},
  pages={24985--25020},
  year={2025},
  doi={10.18653/v1/2025.emnlp-main.1270}
}

@inproceedings{gou2025mind2web2,
  title={{Mind2Web 2}: Evaluating Agentic Search with {Agent-as-a-Judge}},
  author={Gou, Boyu and Huang, Zanming and Ning, Yuting and Gu, Yu and Lin, Michael and Qi, Weijian and Kopanev, Andrei and Yu, Botao and Jim{\'e}nez Guti{\'e}rrez, Bernal and Shu, Yiheng and Song, Chan Hee and Wu, Jiaman and Chen, Shijie and Moussa, Hanane Nour and Zhang, Tianshu and Xie, Jian and Li, Yifei and Xue, Tianci and Liao, Zeyi and Zhang, Kai and Zheng, Boyuan and Cai, Zhaowei and Rozgic, Viktor and Ziyadi, Morteza and Sun, Huan and Su, Yu},
  booktitle={Advances in Neural Information Processing Systems},
  volume={38},
  pages={192393--192451},
  year={2025},
  doi={10.52202/085713-5778}
}

@article{shen2026oneeval,
  title={{One-Eval}: An Agentic System for Automated and Traceable {LLM} Evaluation},
  author={Shen, Chengyu and Hou, Yanheng and Pan, Minghui and He, Runming and Wong, Zhen Hao and Qiang, Meiyi and Liu, Zhou and Liang, Hao and Lai, Peichao and Sheng, Zeang and Zhang, Wentao},
  journal={arXiv preprint arXiv:2603.09821},
  year={2026}
}

@inproceedings{wang2025cigeval,
  title={A Unified Agentic Framework for Evaluating Conditional Image Generation},
  author={Wang, Jifang and Yang, Xue and Wang, Longyue and Xu, Zhenran and Wang, Yiyu and Wang, Yaowei and Luo, Weihua and Zhang, Kaifu and Hu, Baotian and Zhang, Min},
  booktitle={Proceedings of the 63rd Annual Meeting of the Association for Computational Linguistics (Volume 1: Long Papers)},
  pages={12626--12646},
  year={2025},
  doi={10.18653/v1/2025.acl-long.620}
}

@article{chen2026edival,
  title={{EdiVal-Agent}: An Object-Centric Framework for Automated, Fine-Grained Evaluation of Multi-Turn Editing},
  author={Chen, Tianyu and Zhang, Yasi and Zhang, Zhi and Yu, Peiyu and Wang, Shu and Wang, Zhendong and Lin, Kevin and Wang, Xiaofei and Yang, Zhengyuan and Li, Linjie and Lin, Chung-Ching and Xie, Jianwen and Leong, Oscar and Wang, Lijuan and Wu, Ying Nian and Zhou, Mingyuan},
  journal={arXiv preprint arXiv:2509.13399},
  year={2026}
}

@article{zhang2026rewardharness,
  title={{RewardHarness}: Self-Evolving Agentic Post-Training},
  author={Zhang, Yuxuan and Du, Penghui and Li, Bo and Wei, Cong and Miao, Junwen and Zhang, Huaisong and Cai, Songcheng and Wang, Yubo and Jiang, Dongfu and Zhang, Yuyu and Nie, Ping and Chen, Wenhu and Yu, Changqian and Allen, Kelsey R.},
  journal={arXiv preprint arXiv:2605.08703},
  year={2026}
}

@article{song2026vqqa,
  title={{VQQA}: An Agentic Approach for Video Evaluation and Quality Improvement},
  author={Song, Yiwen and Pfister, Tomas and Song, Yale},
  journal={arXiv preprint arXiv:2603.12310},
  year={2026}
}

@article{zeng2026videoargus,
  title={{VideoArgus}: Agentic Rubric-Grounded Unified Evaluation for Video Generation and Editing},
  author={Zeng, Ziyun and Wang, Zixuan and Yu, Yongsheng and Hua, Hang and Luo, Jiebo},
  journal={arXiv preprint arXiv:2608.05485},
  year={2026}
}

@inproceedings{lu2025agentrewardbench,
  title={{AgentRewardBench}: Evaluating Automatic Evaluations of Web Agent Trajectories},
  author={L{\`u}, Xing Han and Kazemnejad, Amirhossein and Meade, Nicholas and Patel, Arkil and Shin, Dongchan and Zambrano, Alejandra and Sta{\'n}czak, Karolina and Shaw, Peter and Pal, Christopher J. and Reddy, Siva},
  booktitle={Second Conference on Language Modeling},
  year={2025},
  url={https://openreview.net/forum?id=fQcUZMPIvu}
}

@inproceedings{shi2026ajbench,
  title={{AJ-Bench}: Benchmarking {Agent-as-a-Judge} for Environment-Aware Evaluation},
  author={Shi, Wentao and Wang, Yu and Zhao, Yuyang and Chen, Yuxin and Feng, Fuli and Hao, Xueyuan and Su, Xi and Gu, Qi and Su, Hui and Cai, Xunliang and He, Xiangnan},
  booktitle={Findings of the Association for Computational Linguistics: ACL 2026},
  pages={25371--25413},
  year={2026},
  doi={10.18653/v1/2026.findings-acl.1269}
}

@article{wang2026reflect,
  title={Time to {REFLECT}: Can We Trust {LLM} Judges for Evidence-based Research Agents?},
  author={Wang, Leyao and He, Yanan and Chen, Peng and Yehudai, Asaf and Liu, Yixin and Ying, Rex and Shmueli-Scheuer, Michal and Cohan, Arman},
  journal={arXiv preprint arXiv:2605.19196},
  year={2026}
}

@inproceedings{zhu2025rigorousagentbenchmarks,
  title={Establishing Best Practices in Building Rigorous Agentic Benchmarks},
  author={Zhu, Yuxuan and Jin, Tengjun and Pruksachatkun, Yada and Zhang, Andy and Liu, Shu and Cui, Sasha and Kapoor, Sayash and Longpre, Shayne and Meng, Kevin and Weiss, Rebecca and Barez, Fazl and Gupta, Rahul and Dhamala, Jwala and Merizian, Jacob and Giulianelli, Mario and Coppock, Harry and Ududec, Cozmin and Kellermann, Antony and Sekhon, Jasjeet and Steinhardt, Jacob and Schwettmann, Sarah and Narayanan, Arvind and Zaharia, Matei A. and Stoica, Ion and Liang, Percy and Kang, Daniel},
  booktitle={Advances in Neural Information Processing Systems},
  volume={38},
  pages={184435--184475},
  year={2025},
  doi={10.52202/085713-5547}
}

@article{barres2025tau2bench,
  title={{$\tau^2$-Bench}: Evaluating Conversational Agents in a Dual-Control Environment},
  author={Barres, Victor and Dong, Honghua and Ray, Soham and Si, Xujie and Narasimhan, Karthik},
  journal={arXiv preprint arXiv:2506.07982},
  year={2025}
}

@inproceedings{xue2025illusion,
  title={An Illusion of Progress? Assessing the Current State of Web Agents},
  author={Xue, Tianci and Qi, Weijian and Shi, Tianneng and Song, Chan Hee and Gou, Boyu and Song, Dawn and Sun, Huan and Su, Yu},
  booktitle={Second Conference on Language Modeling},
  year={2025},
  url={https://openreview.net/forum?id=6jZi4HSs6o}
}

@inproceedings{starace2025paperbench,
  title={{PaperBench}: Evaluating {AI}'s Ability to Replicate {AI} Research},
  author={Starace, Giulio and Jaffe, Oliver and Sherburn, Dane and Aung, James and Chan, Jun Shern and Maksin, Leon and Dias, Rachel and Mays, Evan and Kinsella, Benjamin and Thompson, Wyatt and Heidecke, Johannes and Glaese, Amelia and Patwardhan, Tejal},
  booktitle={Proceedings of the 42nd International Conference on Machine Learning},
  series={Proceedings of Machine Learning Research},
  volume={267},
  pages={56843--56873},
  year={2025}
}

@inproceedings{li2025generationjudgment,
  title={From Generation to Judgment: Opportunities and Challenges of {LLM}-as-a-Judge},
  author={Li, Dawei and Jiang, Bohan and Huang, Liangjie and Beigi, Alimohammad and Zhao, Chengshuai and Tan, Zhen and Bhattacharjee, Amrita and Jiang, Yuxuan and Chen, Canyu and Wu, Tianhao and Shu, Kai and Cheng, Lu and Liu, Huan},
  booktitle={Proceedings of the 2025 Conference on Empirical Methods in Natural Language Processing},
  pages={2757--2791},
  year={2025},
  doi={10.18653/v1/2025.emnlp-main.138}
}

@inproceedings{zou2026unimmmu,
  title={{Uni-MMMU}: A Massive Multi-discipline Multimodal Unified Benchmark},
  author={Zou, Kai and Huang, Ziqi and Dong, Yuhao and Tian, Shulin and Zheng, Dian and Liu, Hongbo and He, Jingwen and Liu, Bin and Qiao, Yu and Liu, Ziwei},
  booktitle={Proceedings of the 64th Annual Meeting of the Association for Computational Linguistics (Volume 1: Long Papers)},
  pages={908--924},
  year={2026},
  doi={10.18653/v1/2026.acl-long.40}
}

@inproceedings{shi2026realunify,
  title={{RealUnify}: Do Unified Models Truly Benefit from Unification? {A} Comprehensive Benchmark},
  author={Shi, Yang and Dong, Yuhao and Ding, Yue and Wang, Yuran and Zhu, Xuanyu and Zhou, Sheng and Liu, Wenting and Tian, Haochen and Wang, Rundong and Wang, Huanqian and Liu, Zuyan and Zeng, Bohan and Chen, Ruizhe and Wang, Qixun and Zhang, Zhuoran and Chen, Xinlong and Tong, Chengzhuo and Li, Bozhou and Liu, Qiang and Wang, Haotian and Yang, Wenjing and Zhang, Yuanxing and Wan, Pengfei and Zhang, Yi-Fan and Liu, Ziwei},
  booktitle={Proceedings of the IEEE/CVF Conference on Computer Vision and Pattern Recognition},
  pages={22488--22497},
  year={2026}
}

@inproceedings{liu2026unison,
  title={{Unison}: Benchmarking Unified Multimodal Models via Synergistic Understanding and Generation},
  author={Liu, Jinyu and Shuai, Xincheng and Ding, Henghui and Jiang, Yu-Gang},
  booktitle={International Conference on Machine Learning (ICML)},
  year={2026}
}

@article{you2026agentasjudge,
  title={{Agent-as-a-Judge}},
  author={You, Runyang and Cai, Hongru and Zhang, Caiqi and Xu, Qiancheng and Liu, Meng and Yu, Tiezheng and Li, Yongqi and Li, Wenjie},
  journal={arXiv preprint arXiv:2601.05111},
  year={2026}
}

@inproceedings{wang2026agenticeval,
  title={{AgenticEval}: Toward Agentic and Self-Evolving Safety Evaluation of Large Language Models},
  author={Wang, Yixu and Wang, Xin and Yao, Yang and Li, Xinyuan and Yang, Xibang and Teng, Yan and Ma, Xingjun and Wang, Yingchun},
  booktitle={Findings of the Association for Computational Linguistics: ACL 2026},
  pages={14789--14808},
  year={2026},
  doi={10.18653/v1/2026.findings-acl.727}
}

\end{document}